%% file: neurips_2026.tex
\documentclass{article}

 \usepackage[preprint]{neurips_2026}

\usepackage[utf8]{inputenc} % allow utf-8 input
\usepackage[T1]{fontenc}    % use 8-bit T1 fonts
\usepackage{hyperref}       % hyperlinks
\usepackage{url}            % simple URL typesetting
\usepackage{booktabs}       % professional-quality tables
\usepackage{amsfonts}       % blackboard math symbols
\usepackage{nicefrac}       % compact symbols for 1/2, etc.
\usepackage{microtype}      % microtypography
\usepackage{xcolor}         % colors
\usepackage{xspace}
\usepackage{graphicx}
\usepackage{adjustbox}
\usepackage{amsmath}
\usepackage{seqsplit}

\usepackage[most]{tcolorbox}
\usepackage{fontawesome5}

\usepackage{pifont} % Required for \ding commands
\newcommand{\cmark}{\textcolor{green}{\ding{51}}}
\newcommand{\xmark}{\textcolor{red}{\ding{55}}}

\newcommand{\mypara}[1]{\noindent\textbf{#1}}

\tcbset{
  panel/.style={
    enhanced,breakable,boxrule=0.7pt,arc=2.2mm,
    left=3.5mm,right=3.5mm,top=1.8mm,bottom=1.8mm,
    before skip=1mm,after skip=1mm,
  }
}

\newtcolorbox{titlebox}{panel,colback=titlebg,colframe=titlebg!85!black,arc=2.2mm,top=2.2mm,bottom=2.2mm}
\newtcolorbox{userbox}{panel,colback=userbg,colframe=userframe}
\newtcolorbox{agentbox}{panel,colback=agentbg,colframe=agentframe}
\newtcolorbox{toolbox}{panel,colback=toolbg,colframe=toolframe}
\newtcolorbox{responsebox}{panel,colback=respbg,colframe=respframe}

\newtcolorbox{innercode}{
  enhanced,boxrule=0.6pt,arc=1.5mm,
  colback=codebg,colframe=codeframe,
  left=2.5mm,right=2.5mm,top=1mm,bottom=1mm,
  fontupper=\ttfamily\small\color{softtext},
}

\usepackage{enumitem}     % [nosep,leftmargin=*] 
\usepackage{float}

\newcommand{\bench}{\texttt{PhysicianBench}\xspace}

\title{
PhysicianBench: Evaluating LLM Agents in Real-World EHR Environments
}

\author{
    Ruoqi Liu,
    Imran Q. Mohiuddin,
    Austin J. Schoeffler,\\
    \textbf{Kavita Renduchintala,}
    \textbf{Ashwin Nayak,}
    \textbf{Prasantha L.~Vemu,}\\
    \textbf{Shivam C.~Vedak,}
    \textbf{Kameron C.~Black,}
    \textbf{John L.~Havlik,}\\
    \textbf{Isaac Ogunmola,}
    \textbf{Stephen P.~Ma,}
    \textbf{Roopa Dhatt,}
    \textbf{Jonathan H.~Chen} \\ \\
  Stanford University \\
  \texttt{\{ruoqiliu, jonc101\}@stanford.edu}
}

\begin{document}

\maketitle

\vspace{-1.0cm}
\begin{center}
    \faGlobe: \url{https://healthrex.github.io/PhysicianBench/ }
\end{center}
\vspace{0cm}

\input{0_abstract}
\input{1_intro}

\input{5_relatedwork}
\input{2_method}
\input{3_results}
\input{4_discussion}
\input{7_acknowledgement}

\bibliographystyle{plain}

\bibliography{ref}

%%%%%%%%%%%%%%%%%%%%%%%%%%%%%%%%%%%%%%%%%%%%%%%%%%%%%%%%%%%%

\input{9_appendix}

%%%%%%%%%%%%%%%%%%%%%%%%%%%%%%%%%%%%%%%%%%%%%%%%%%%%%%%%%%%%

% \newpage
% \input{checklist.tex}

\end{document}

%% file: 0_abstract.tex
\begin{abstract}
We introduce \bench, a benchmark for evaluating LLM agents on physician tasks grounded in real clinical setting within electronic health record (EHR) environments. Existing medical agent benchmarks primarily focus on static knowledge recall, single-step atomic actions, or action intent without verifiable execution against the environment. As a result, they fail to capture the long-horizon, composite workflows that characterize real clinical systems.
\bench comprises 100 long-horizon tasks adapted from real consultation cases between primary care and subspecialty physicians, with each task independently reviewed by a separate panel of physicians. Tasks are instantiated in an EHR environment with real patient records and accessed through the same standard APIs used by commercial EHR vendors. Tasks span 21 specialties (e.g., cardiology, endocrinology, oncology, psychiatry) and diverse workflow types (e.g., diagnosis interpretation, medication prescribing, treatment planning), requiring an average of 27 tool calls per task. Solving each task requires retrieving data across encounters, reasoning over heterogeneous clinical information, executing consequential clinical actions, and producing clinical documentation. Each task is decomposed into structured checkpoints (670 in total across the benchmark) capturing distinct stages of completion graded by task-specific scripts with execution-grounded verification.
Across 12 proprietary and open-source LLM agents, the best-performing model achieves only 46\% success rate (pass@1), while open-source models reach at most 19\%, revealing a substantial gap between current agent capabilities and the demands of real-world clinical workflows.
\bench provides a realistic and execution-grounded benchmark for measuring progress toward autonomous clinical agents.
\end{abstract}

\begin{figure}[H]
    \centering
\vspace{-8pt}
\includegraphics[width=\linewidth]{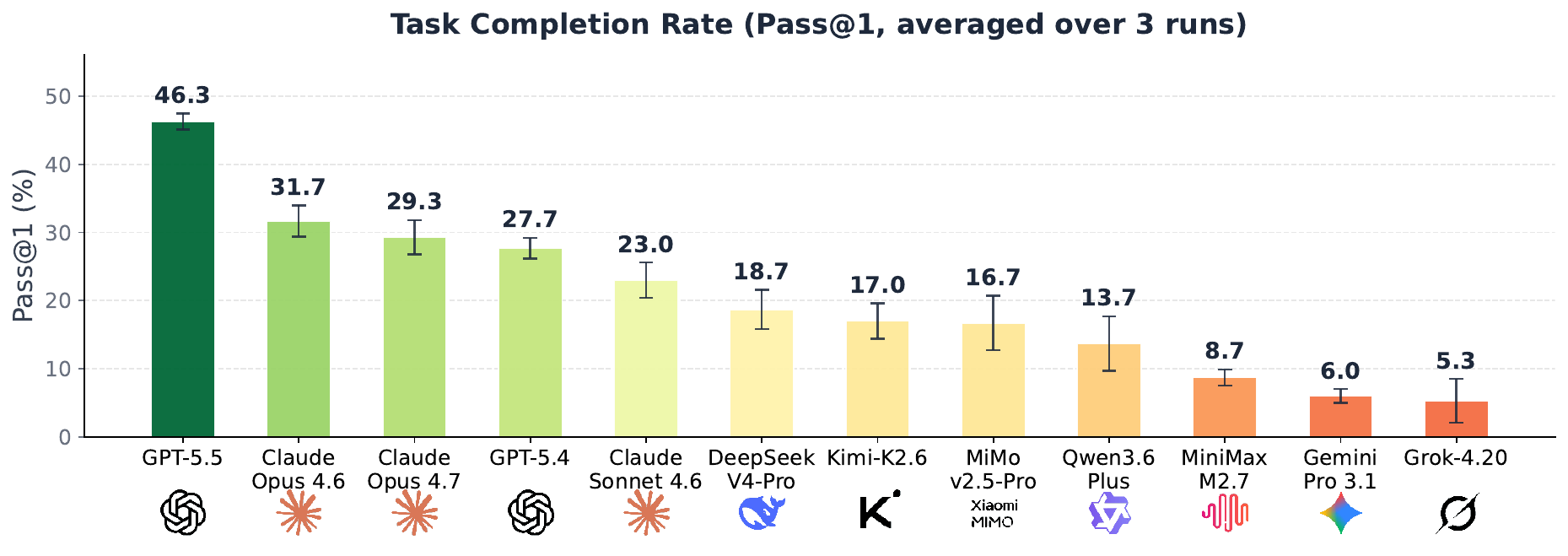}
\vspace{-16pt}
    \caption{\bench overall model performance ranked by success rate (pass@1).}
    \label{fig:banner}
\vspace{-12pt}
\end{figure}

%% file: 1_intro.tex
\section{Introduction}
\label{sec:intro}

\begin{figure}[!t]
    \centering
\includegraphics[width=\linewidth]{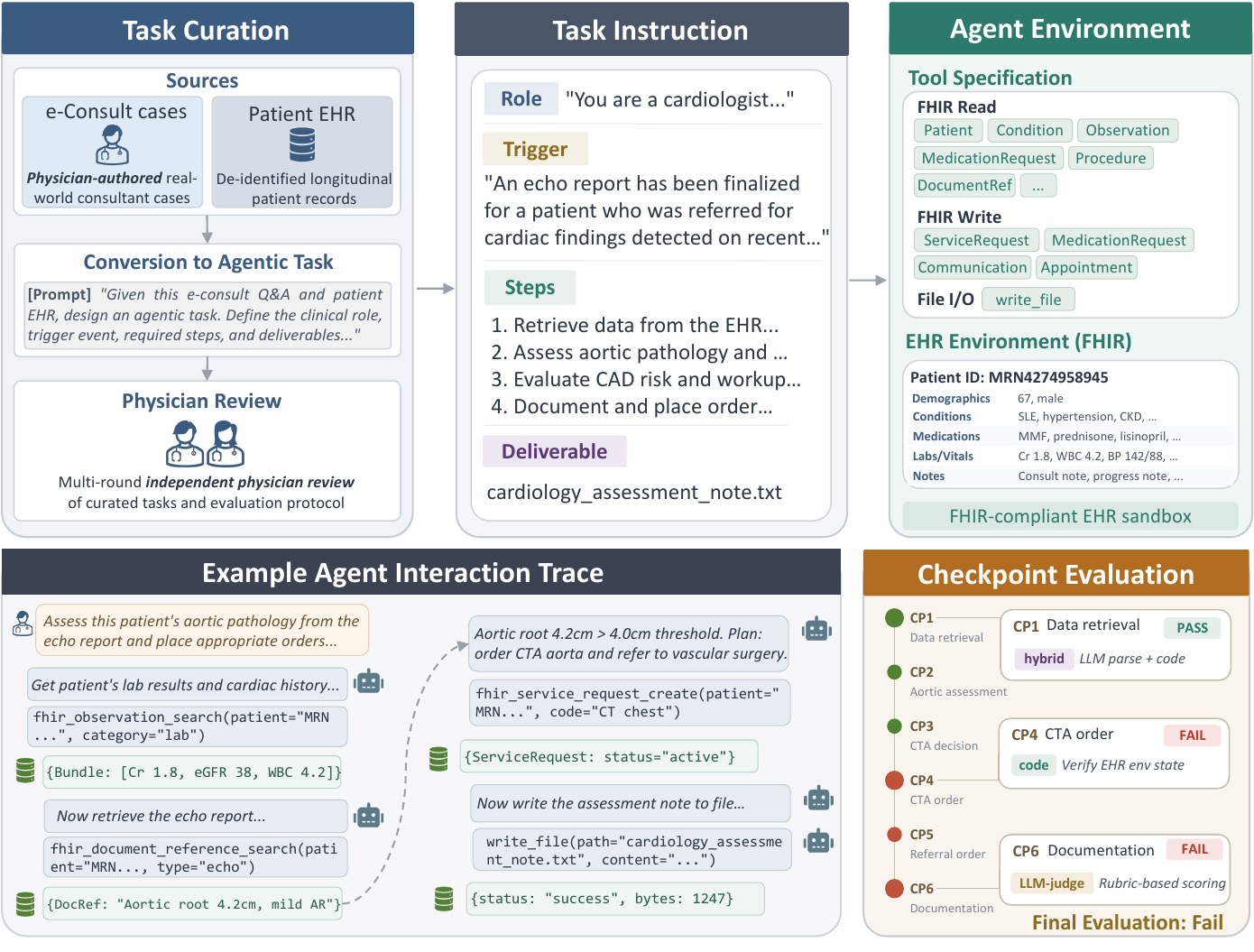}
\vspace{-8pt}
    \caption{Overview of \bench. Tasks are curated from real consultation cases, paired with real EHR environments, and further validated by physicians independently. Each task instruction specifies a clinical role, trigger event, required steps, and expected deliverable. During execution, the agent interacts with EHR environment through multi-step tool calls. Agent performance is assessed via a sequence of checkpoints for each step of the task.}
    \label{fig:overview}
    \vspace{-10pt}
\end{figure}

Large language models (LLMs) have demonstrated strong clinical knowledge and are increasingly deployed in health contexts to support information access, clinical decision-making, and patient care~\citep{singhal2023large,korom2025ai}.
A more consequential shift is now underway: beyond answering questions, LLMs are being instantiated as \emph{agents} that can autonomously plan, use tools, interact with external systems, and execute multi-step actions~\citep{yao2022react,xu2025theagentcompany,yao2025tau}.
In healthcare, this shift is particularly impactful.
Real clinical work requires retrieving information from electronic health records (EHRs), reasoning over heterogeneous data sources, coordinating with clinical systems, and executing consequential actions such as placing orders, prescribing medications, and documenting clinical plans.
LLM agents that reliably perform such work could alleviate the well-documented EHR burden that now dominates clinicians' workdays and contributes to professional burnout~\citep{sinsky2016allocation,holmgren2024national}.
% Conversely, unreliable agent behavior in clinical settings carries real risk of patient harm~\citep{ramaswamy2026chatgpt}.

Realizing this potential, and measuring progress toward it, requires benchmarks that test not only what agents know, but whether they can reliably complete clinical workflows.
Existing medical benchmarks fall short of this goal in three ways.
First, they rarely evaluate long-horizon, multi-step agent behavior, instead focusing on question answering~\citep{jin2021medqa,jin2019pubmedqa}, clinical conversation and response quality~\citep{arora2025healthbench,schmidgall2024agentclinic}, or single-step atomic actions~\citep{jiang2025medagentbench}.
Second, they provide limited evaluation in real EHR environments: benchmarks may assess the intent of an action~\citep{jiang2025medagentbench}, rather than requiring agents to execute actions in an EHR system and verifying that those actions correctly update the underlying clinical state.
Third, their tasks are often derived from exams~\citep{jin2021medqa} or synthetic patient simulators~\citep{schmidgall2024agentclinic}, rather than real clinical cases that are validated by physicians.

In this work, we introduce \bench (Fig.~\ref{fig:overview}), a benchmark for evaluating LLM agents on long-horizon tasks in a real EHR environment.
\bench comprises 100 tasks curated from real clinical consultation cases between primary care and subspecialty physicians, spanning 21 specialties and diverse workflows.
Each task requires an agent to retrieve patient data, reason across encounters, execute clinical actions, and produce clinical documentation, averaging 27 tool calls per task.
Tasks run against EHR instances accessed through standard Fast Healthcare Interoperability Resources (FHIR) APIs, a widely used standard for API-based interoperability in modern EHR systems~\citep{hl7fhir}.
Task correctness is graded via task-specific checkpoints that capture intermediate reasoning steps and final action outcomes, combining execution-grounded verification of the resulting FHIR state with complementary LLM-based judgments.
Each task is independently reviewed by a panel of physicians through a multi-round annotation pipeline to ensure clinical fidelity.

We evaluate 12 proprietary and open-source LLMs. The best-performing agent, GPT-5.5, completes only 46\% of tasks (pass@1), with Claude Opus 4.6 and 4.7 reaching 32\% and 29\% respectively, while the strongest open-source model (DeepSeek V4-Pro) reaches 19\%.
% The same Claude Opus family that nearly saturates MedAgentBench's atomic tasks achieves less than a third of \bench, quantifying the gap between atomic skill competence and coherent workflow execution.
Reliability is also limited: pass\textasciicircum{}3 across three independent runs reaches only 28\% even for the best model, indicating that successful task completion is often stochastic.
Error analysis further reveals that failures are distributed across retrieval, reasoning, and action execution rather than concentrated in any single bottleneck.

Our contributions are as follows:
\begin{itemize}[leftmargin=*]
   \item \textbf{A real-world, physician-validated benchmark.} \bench provides 100 long-horizon clinical tasks sourced from real consultation cases, both authored and independently reviewed by physicians, instantiated in a FHIR-compliant EHR environment.
    \item \textbf{Execution-grounded and verifiable evaluation.} Agent actions are verified against the post-execution EHR state to enable verifiable execution, while structured per-task checkpoints support effective diagnostics of agent failure modes.
    \item \textbf{Measuring progress of agents on real clinical tasks.} An evaluation of 12 LLM agents quantifies how far current agents remain from autonomously and reliably handling real-world clinical tasks.
    \item \textbf{Fully open-source release.} To support reproducible evaluation and future development of agents, we release the benchmark with environments, the agent framework, and the evaluation harness.
\end{itemize}

%% file: 5_relatedwork.tex
\section{Related Work}
\label{sec:related-work}

\paragraph{General-domain agent benchmarks.}
Evaluating LLMs as autonomous agents has progressed rapidly from narrow skill tests to real, long-horizon simulations. Early efforts established multi-environment evaluation~\citep{liu2023agentbench} and function-calling accuracy~\citep{patil2025berkeley}, followed by domain-specific benchmarks such as software engineering~\citep{jimenez2024swebench} and web navigation~\citep{zhou2024webarena}.

More recent benchmarks emphasize multi-step tasks in real environments: $\tau$-bench~\citep{yao2025tau} evaluates tool-agent-user interaction in retail and airline domains; TheAgentCompany~\citep{xu2025theagentcompany} simulates a software company spanning web browsing and code execution; Toolathlon~\citep{li2025toolathlon} scales to 32 applications and 604 tools; and MCP-Bench~\citep{wang2025mcp} connects LLMs to 28 live MCP servers spanning 250 tools across finance, travel, and scientific computing. These benchmarks demonstrate that multi-step, consequential task execution remains an open challenge. However, none address healthcare, a domain with unique requirements including specialized clinical terminology, complex multi-system reasoning, and regulatory data standards such as FHIR.

\paragraph{Medical agent benchmarks.}
Traditional medical AI benchmarks such as MedQA~\citep{jin2021medqa}, PubMedQA~\citep{jin2019pubmedqa}, and MedMCQA~\citep{pal2022medmcqa} evaluate clinical knowledge recall through static multiple-choice questions derived from licensing examinations, without involving tool use, EHR interaction, or action execution. More recently, HealthBench~\citep{arora2025healthbench} expands clinical evaluation to 5,000 open-ended health conversations graded by physician-designed rubrics, and its extension HealthBench Professional~\citep{hicks2026healthbench} further stratifies these scenarios across 27 clinical specialties. MedHELM~\citep{bedi2026holistic} aggregates a broad set of clinical NLP tasks into a unified holistic evaluation suite. These efforts substantially broaden the scope of medical LLM evaluation, but still grade static model responses without tool use, EHR interaction, or action execution.

AgentClinic~\citep{schmidgall2024agentclinic} moves toward interactive evaluation by simulating multi-turn patient-doctor conversations, demonstrating that diagnostic accuracy can drop to one-tenth of static performance.
However, it remains diagnosis-centric and does not involve EHR systems, structured data manipulation, or clinical action execution such as ordering medications or placing referrals. EHRAgent~\citep{shi2024ehragent} equips LLMs with code-generation interfaces to translate clinical questions into SQL queries over EHR data warehouses, but operates on site-specific schemas rather than interoperable standards. FHIR-AgentBench~\citep{lee2025fhir} grounds 2,931 clinical questions in FHIR-structured data, evaluating retrieval precision and answer correctness, but focuses exclusively on question answering without workflow execution or action creation. AgentEHR~\citep{liao2026agentehr} challenges agents to perform diagnosis and treatment planning through long-range interactive reasoning over raw EHR databases, but does not involve a FHIR-compliant environment or structured action execution (e.g., creating orders or referrals).

The most closely related work is MedAgentBench~\citep{jiang2025medagentbench}, which also provides a FHIR-compliant interactive environment with 300 clinician-authored tasks across 10 templates. It represents a significant advance in grounding medical agent evaluation in realistic EHR infrastructure. However, its tasks are predominantly single-step atomic operations, such as retrieving a single lab value or placing a single order, evaluating individual clinical skills in isolation rather than the composite, multi-step workflows that characterize real clinical practice.

\bench addresses these limitations by combining three properties absent from any single prior benchmark:
(1)~long-horizon, multi-step agent tasks reflecting the real clinical workflow from information retrieval through reasoning, action execution, and documentation;
(2)~a real EHR environment accessed through standard FHIR APIs, with execution-grounded verification of actions against the resulting EHR state; and
(3)~tasks grounded in real clinical consultation cases independently validated by a separate panel of physicians.
Table~\ref{tab:benchmark-comparison} summarizes the key differences between \bench and existing benchmarks along these dimensions.

\begin{table}[!t]
\centering
\caption{Comparison of \bench with related benchmarks. ``EHR Env.'' indicates whether the benchmark provides an EHR system with production-standard APIs rather than static EHR data dumps; ``Verifiable Execution'' indicates whether the evaluation verifies the resulting EHR state after execution; ``Specialties'' reports the number of clinical specialties covered (``--'' indicates not reported by the original paper); ``Clinician-validated'' indicates whether tasks are validated by clinicians.}
\label{tab:benchmark-comparison}
\adjustbox{max width=\linewidth}{
\begin{tabular}{lccccc}
\toprule
Benchmark & EHR Env. & Long-horizon & Verifiable Execution & Specialties & Clinician-validated \\
\midrule
HealthBench Professional~\citep{hicks2026healthbench}      & \xmark & \xmark & \xmark & 27 & \cmark \\
AgentClinic~\citep{schmidgall2024agentclinic}  & \xmark & \xmark & \xmark & 9  & \xmark \\
EHRAgent~\citep{shi2024ehragent}               & \xmark & \xmark & \xmark & -- & \xmark \\
MedAgentBench~\citep{jiang2025medagentbench}   & \cmark & \xmark & \xmark & -- & \cmark \\
FHIR-AgentBench~\citep{lee2025fhir}            & \cmark & \xmark & \xmark & -- & \xmark \\
AgentEHR~\citep{liao2026agentehr}              & \xmark & \cmark & \xmark & -- & \xmark \\
\midrule
\bench                                  & \cmark & \cmark & \cmark & 21 & \cmark \\
\bottomrule
\end{tabular}}
\end{table}

%% file: 2_method.tex
\section{Benchmark Environment and Evaluation}
\label{sec:benchmark}
% ============================================================================
\bench is an integrated benchmark system in which an agent receives a clinical task instruction, interacts with a FHIR-compliant EHR through structured tool calls, and is evaluated against a set of clinically grounded checkpoints.
Figure~\ref{fig:overview} illustrates the end-to-end pipeline.
We first formalize the task structure~(\S\ref{sec:task-formulation}), then describe the EHR environment and available tools~(\S\ref{sec:environment}), and finally present the checkpoint-based evaluation framework~(\S\ref{sec:evaluation}).

% ----------------------------------------------------------------------------
\subsection{Task Formulation}
\label{sec:task-formulation}
% ----------------------------------------------------------------------------

We formulate each task as a partially observable Markov decision process (POMDP) $(\mathcal{S}, \mathcal{A}, T, \mathcal{O}, \Omega, I, C)$.
The state space $\mathcal{S}$ comprises the full EHR contents, including patient records, agent-created resources, and workspace files.
The action space $\mathcal{A}$ is defined by 14 structured tool calls (\S\ref{sec:environment}).
The transition function $T$ is deterministic: write actions (e.g., creating a \texttt{MedicationRequest}) modify the EHR state, while read actions (e.g., searching for lab results) leave it unchanged.
The observation space $\mathcal{O}$ consists of the FHIR JSON bundles returned by search tools, confirmation responses from write tools, and file contents.
The observation function $\Omega$ governs what the agent perceives: it cannot observe the full EHR state but must actively query relevant resources to build a working clinical picture, mirroring how clinicians navigate real EHR systems.
Each task instance additionally specifies a natural-language instruction $I$, which defines the clinical scenario (the agent's role, patient identifier, current date, multi-step requirements, and expected deliverables), and a set of evaluation checkpoints $C = \{c_1, \dots, c_K\}$ used for grading (\S\ref{sec:evaluation}).

At each turn $t$, the agent selects a tool call $a_t \in \mathcal{A}$ conditioned on $I$ and its observation history $(o_1, \dots, o_{t-1})$, receives an observation $o_t$, and the environment transitions to $s_{t+1} = T(s_t, a_t)$.
The interaction terminates when the agent issues a final response or reaches the maximum step limit.

% Each task is designed as a composite task requiring atomic capabilities spanning: (1) data retrieval (e.g., querying labs, reviewing clinical notes); (2) clinical reasoning (e.g., risk stratification, differential diagnosis); (3) action execution (e.g., placing medication orders, initiating referrals); (4) clinical documentation (e.g., ).
% This composite design ensures that no single capability suffices to complete a task; we detail the task construction process and clinical coverage in Section~\ref{sec:curation}.

% ----------------------------------------------------------------------------
\subsection{EHR Environment and Agent Tools}
\label{sec:environment}
% ----------------------------------------------------------------------------

\mypara{EHR server.}
FHIR~\citep{hl7fhir} is the standard for health information exchange, supported by major commercial EHR vendors (e.g., Epic, Cerner, Oracle Health).
We build our environment on FHIR R4 to ensure that agent tool calls, data representations, and clinical workflows closely mirror those in production healthcare systems.
Concretely, we use the open-source HAPI FHIR JPA server~\citep{hapifhir} backed by a persistent H2 database.
Each task instance runs in an isolated Docker container hosting a dedicated FHIR server pre-loaded with the corresponding patient record encoded using standardized clinical terminologies (e.g., LOINC codes for laboratory observations and vitals).
% Patient data is imported via structured POST requests that create FHIR resources with appropriate coding systems and reference links.

\mypara{Agent tools.}
The agent interacts with the EHR through a set of tools including (1) FHIR \texttt{GET} requests to query patient demographics, the problem list (Condition), laboratory results and vital signs (Observation), social history, medication orders (MedicationRequest), completed procedures (Procedure), clinical notes (DocumentReference), and existing service requests (ServiceRequest); (2) FHIR \texttt{POST} requests to create medication orders, service requests for diagnostic studies and referrals, appointments, and patient communications; and (3) Utility tool allowing the agent to write output files to the workspace.
All tools accept structured parameters following the function-calling schema; search tools return JSON bundles, while write tools return the created resource or a file-write confirmation.
% Importantly, resources created by write operations persist in the EHR and are subsequently verifiable by the evaluation harness (\S\ref{sec:evaluation}).
The full tool inventory, with FHIR resource types and key arguments, is provided in Table~\ref{tab:tool_inventory} (Appendix~\ref{app:agent-framework}).

\mypara{Agent Framework.}
We implement a lightweight tool-calling agent framework to enable fair comparison across models with different native agent integrations.
At each turn, the model receives the task instruction and complete interaction history, may produce reasoning tokens, and emits one or more tool calls through an OpenAI-compatible function-calling interface.
The framework dispatches each call, appends the resulting observation to the history, and iterates until the agent returns a final response, reaches the maximum step limit, or triggers an early-termination safeguard.
Beyond this core loop, the framework includes robustness mechanisms for tool-error handling, overlong-output truncation, loop and stuck-state detection.
All model responses, tool calls, and tool observations are recorded in a structured trajectory log for failure analysis.
Full implementation details, configuration settings, and safeguard thresholds are provided in Appendix~\ref{app:agent-framework}.

% ----------------------------------------------------------------------------
\subsection{Checkpoint Evaluation}
\label{sec:evaluation}
% ----------------------------------------------------------------------------

Rather than evaluating only the final task output, we decompose each task into a set of checkpoints $C = \{c_1, \dots, c_K\}$, where each checkpoint represents a discrete, clinically meaningful milestone.
\bench contains 670 checkpoints in total across the 100 tasks, enabling fine-grained diagnostics that identify which stages of clinical workflows are most challenging for current agents.

\mypara{Checkpoint taxonomy.}
Checkpoints follow a consistent progression that mirrors the real clinical decision-making.
Each checkpoint is assigned a type reflecting the capability it evaluates:
\begin{itemize}[nosep,leftmargin=*]
    \item \textbf{Data retrieval}: verifies that the agent queried the necessary resources (e.g., demographics, laboratory results, conditions) and surfaced the critical data points required for downstream reasoning.
    \item \textbf{Clinical reasoning}: evaluates the agent's interpretation of retrieved data across four subtypes: calculating clinical scores from patient data (e.g., CHA$_2$DS$_2$-VASc from demographics and coded diagnoses), interpreting and synthesizing clinical findings (e.g., recognizing eGFR discrepancies between creatinine-based and cystatin~C-based estimates), making clinical judgments such as differential diagnosis, treatment selection, or medication management (e.g., anticoagulation strategy for low-risk atrial fibrillation), and applying clinical criteria to patient data (e.g., ruling out proximal tubular dysfunction based on urinalysis findings).
    \item \textbf{Action execution}: validates that the agent created appropriate FHIR resources, including \texttt{MedicationRequest} for prescriptions (e.g., high-intensity statin with correct dose range) and \texttt{ServiceRequest} for diagnostic studies and referrals (e.g., CT angiography, HFE gene testing).
    \item \textbf{Documentation}: assesses the completeness and clinical soundness of the agent's written output (e.g., assessment notes, management plans), including consistency with retrieved data, adherence to clinical guidelines, and absence of clinically unsafe recommendations.
\end{itemize}

\mypara{Grader types.}
Each checkpoint is graded by one of three grader types, selected based on the nature of the criterion: (1) code graders query the FHIR server to verify that the agent created the expected resources (e.g., a \texttt{MedicationRequest} matching the required drug, dose, and frequency); (2) hybrid graders combine deterministic ground-truth computation from FHIR data with LLM-assisted extraction from the agent's free-text output (e.g., comparing a CHA$_2$DS$_2$-VASc score the agent reports against the score computed from coded diagnoses); and (3) LLM-judge graders score clinical reasoning and documentation against a structured rubric with three-level outcomes (pass/partial/fail). Full grader-type specifications and example rubrics are provided in Appendix~\ref{app:graders}.
% [FIGURE/TABLE placeholder]: Distribution of grader types across the
% benchmark; example checkpoint sequence for a representative task.

\begin{figure}[!t]
    \centering
    \includegraphics[width=\linewidth]{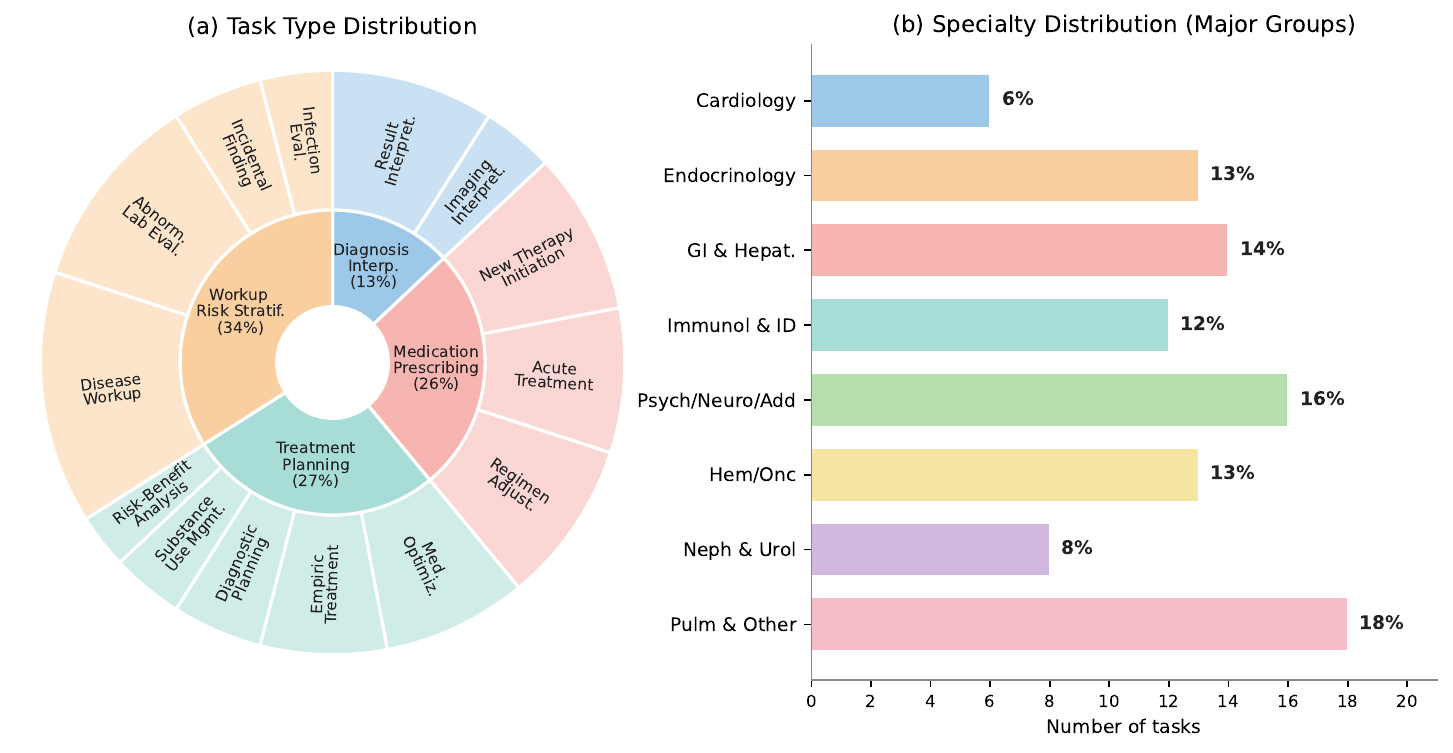}
    \caption{Distribution of tasks in \bench. (a)~Task type distribution as a two-level sunburst: the inner ring shows four high-level clinical workflow types, and the outer ring shows finer-grained subtypes. (b)~Distribution across eight major groups with full list in Table~\ref{tab:task_taxonomy_by_specialty}.}
    \label{fig:task_distribution}
\end{figure}
% ============================================================================
\section{Task Design and Curation}
\label{sec:curation}
% ============================================================================

We now describe how tasks in \bench were constructed and validated.
Each task is grounded in a real clinical case and undergoes multi-round physician review before inclusion in the benchmark.
Figure~\ref{fig:annotation_pipeline} summarizes the curation pipeline.

% ----------------------------------------------------------------------------
\subsection{Task Sourcing and Construction}
\label{sec:task-construction}
% ----------------------------------------------------------------------------

\mypara{Source material.}
Tasks are adapted from real electronic consultation (e-consult) cases drawn from clinical practice. An e-consult is an asynchronous, EHR-based specialty consultation: a primary care physician (PCP) submits a clinical question to a subspecialist, who reviews the patient's full EHR and returns a written assessment with recommended workup, treatment, or referral for the PCP to act on. During task curation, we intentionally diversify the framing beyond the e-consult itself, recasting cases as general clinical tasks across varied roles and settings so the agent's workflow is not limited to static consultation. These cases naturally embody the complex, long-horizon, multi-step clinical reasoning that \bench aims to evaluate.

\mypara{Patient records.}
The patient data underlying each task is derived from real clinical records in the STAnford Research Repository (STARR)~\citep{datta2020new}, a de-identified clinical data warehouse linked to Stanford Health Care EHR systems.
To further protect patient privacy while preserving clinical fidelity, we apply additional perturbations to the source records, including date shifts, demographic modifications, and adjustments to clinical values that preserve the original clinical meaning, which are verified as part of the physician validation pipeline (\S\ref{sec:validation}). 
This approach grounds tasks in realistic clinical complexity, capturing the co-morbidity patterns, data irregularities, and longitudinal health trajectories of real patient populations that fully synthetic records typically lack.

\mypara{Task construction.}
From each source case, we construct a task instance comprising four artifacts:
(a)~a natural-language instruction specifying the clinical context (date, physician role, patient background), multi-step task requirements, and expected deliverables;
(b)~the corresponding patient record encoded in FHIR and mapped to standardized terminologies;
(c)~an evaluation test suite implementing checkpoint definitions and grader logic; and
(d)~a reference solution summary and review checklist supporting structured physician validation.

% ----------------------------------------------------------------------------
\subsection{Clinical Coverage}
\label{sec:coverage}
% ----------------------------------------------------------------------------

\bench comprises 100 physician-validated agent tasks spanning diverse clinical workflows and subspecialties (Figure~\ref{fig:task_distribution}). Tasks are organized into four high-level clinical workflow types (Figure~\ref{fig:task_distribution}a), each further divided into finer-grained subtypes. For example, Workup \& Risk Stratification consists of disease workup, abnormal lab evaluation, incidental finding followup, and infection evaluation. Specifically, these tasks are drawn from 21 clinical subspecialties grouped into 8 major groups (Figure~\ref{fig:task_distribution}b). Appendix~\ref{app:taxonomy} provides the detailed statistics of the clinical coverage.

% ----------------------------------------------------------------------------

\subsection{Physician Validation}
\label{sec:validation}
% ----------------------------------------------------------------------------
To ensure clinical fidelity, every task undergoes multi-round review by a panel of 11 human physicians (background detailed in Appendix~\ref{app:annotators}) through an interactive annotation pipeline. Initial task artifacts and patient EHRs, together with a review checklist tailored to the task's content, then enter the validation pipeline (Figure~\ref{fig:annotation_pipeline}, Appendix~\ref{app:annotation}), which proceeds through three stages:

\textbf{(1) Initial review.}
A domain expert reviews the task against the structured checklist (Figure~\ref{fig:annotation_ui_initial}), which targets four dimensions:
instruction clarity (whether the task is unambiguous and actionable for an agent),
clinical accuracy (whether reasoning chains and clinical recommendations are medically sound),
checkpoint validity (whether evaluation criteria are appropriate and complete), and
EHR consistency (whether referenced clinical data matches the patient record).
The reviewer provides categorical assessments and free-text comments for each checklist item.

\textbf{(2) Agentic revision.}
An LLM agent parses the reviewer's structured assessments and free-text comments and produces targeted edits across all task artifacts. Each edit is recorded as a structured diff capturing the source review item, the affected file and location, the before/after text, and a rationale, so that changes can be individually approved or rejected in stage~(3). When a change requires consistency edits to other artifacts, the agent emits these as auto-synchronized companion diffs grouped under the primary change.

\textbf{(3) Approval review.}
Physicians review each proposed change in a diff-based interface (Figure~\ref{fig:annotation_ui_diff}), seeing the before-and-after text for every edit, and approve or reject individual changes with rationale.
Rejected changes trigger another revision cycle: stages (2) and (3) repeat until all changes are approved.
Tasks reaching full approval are marked as finalized and included in the benchmark. This iterative human-in-the-loop process ensures that the final benchmark tasks meet clinical standards for accuracy, clarity, and safety.

%% file: 3_results.tex
\section{Experiments}
\label{sec:experiments}

\subsection{Experimental Setup}
\label{sec:setup}

\mypara{Models.}
We evaluate multiple LLMs spanning both proprietary and open-source families.
Proprietary models include GPT-5.4~\citep{openai_gpt54_2026}, GPT-5.5~\citep{openai_gpt55_2026}, Claude Opus~4.6~\citep{anthropic_opus46_2026}, Claude Sonnet~4.6~\citep{anthropic_sonnet46_2026} and Claude Opus~4.7~\citep{anthropic_opus47_2026}, Qwen3.6-Plus~\citep{qwen36plus_2026}, MiniMax M2.7~\citep{minimax_m27_2026}, MiMo-v2.5-Pro~\citep{xiaomi_mimo_v25_pro_2026}, Gemini Pro~3.1~\citep{google_gemini31pro_2026}, and Grok-4.20~\citep{xai_grok420_2026}.
Open-source models include DeepSeek V4-Pro~\citep{deepseek2026v4}, and Kimi-K2.6~\citep{moonshot_kimi_k26_2026}.
All models use the same minimal tool-calling agent loop described in \S\ref{sec:task-formulation}, with access to the full set of FHIR tools (\S\ref{sec:environment}).
We allow up to 100 tool-calling turns per task and use the provider's default temperature. For models that support a reasoning-effort parameter, we set it to \texttt{high} and run 3 independent trials to compute the reliability metrics.

\mypara{Metrics.}
We report three complementary metrics.
\textbf{Pass@1} measures the fraction of tasks an agent fully completes in a single attempt, reflecting practical deployment performance.
\textbf{Pass@3} estimates the probability that at least one of $k{=}3$ independent runs succeeds, following the unbiased estimator of \citep{chen2021evaluating}: $\text{pass@}k = 1 - \binom{n-c}{k} / \binom{n}{k}$, where $n$ is the number of runs and $c$ is the number of successes.
This captures whether a model can solve a task when given multiple attempts.
Conversely, \textbf{Pass\textasciicircum{}3} \citep{yao2025tau} $= c^k / n^k$ estimates the probability that all $k$ runs succeed, measuring consistency and reliability.
% A large gap between pass@3 and pass\textasciicircum{}3 indicates stochastic failures amenable to retry strategies, whereas a small gap signals systematic inability.
We also report the average number of tool-calling turns (\textbf{\#Turns}) as a measure of agent efficiency.

\mypara{Environment.}
Each task ships as a self-contained Docker image bundling a HAPI FHIR JPA server~\citep{hapifhir} with the patient record pre-loaded into a persistent H2 database, ensuring exact reproducibility without any runtime data ingestion.
A fresh container is provisioned per task and torn down after evaluation, so no state leaks across evaluations.

\subsection{Main Results}
\label{sec:main-results}

Table~\ref{tab:main_results} reports comparison results for all models.
GPT-5.5 leads by a wide margin, achieving 46.3\% pass@1, an absolute gap of 14.6 points over the next-best model Claude Opus~4.6 (31.7\%). The strongest open-source model, DeepSeek V4-Pro, reaches 18.7\%, surpassing several proprietary models.
GPT-5.5's lead extends to reliability: its Pass\textasciicircum{}3 of 28.0\% is roughly 10 absolute points above the next-best (Claude Opus~4.6 and Claude Opus~4.7, both 18.0\%) and nearly an order of magnitude above weaker models (Grok-4.20, MiniMax M2.7, all at 1.0\%). Even so, GPT-5.5 falls well short of the bar required for autonomous clinical use, with fewer than one in three tasks solved consistently across three independent runs, and the remaining models cluster below 32\% pass@1, leaving the benchmark far from saturated.

\input{tables/results_main}
\input{tables/results_speciality_breakdown}

\mypara{Variation across specialties.}
Table~\ref{tab:specialty_breakdown} reports pass@1 broken down by clinical specialty.
GPT-5.5 leads in seven of the eight specialty groups; the only exception is Nephrology/Urology, where the Claude family tops the chart at 33.3\% and GPT-5.5 reaches 29.2\%.
Within each group, performance spans a wide range (close to zero for the weakest models, approaching 60\% for GPT-5.5 in Endocrinology), indicating that specialty-specific clinical priors and reasoning patterns differ substantially across models.

\mypara{Variation across task types.}
Performance varies systematically by task type (Table~\ref{tab:task_type_breakdown} in Appendix~\ref{app:results}).
Treatment Planning is consistently the hardest task type, reflecting its open-ended, multi-step nature: the best non-GPT-5.5 model reaches only 23.5\% on Treatment Planning compared with 43.6\% on Diagnosis. GPT-5.5's highest score falls in Workup \& Risk Stratification (54.9\%), opening a 17-point gap over Claude Opus~4.6 on that column and suggesting its largest gains come from longer-horizon, multi-source workup tasks.

\input{tables/results_failure_distribution}

\subsection{Error Analysis}
\label{sec:error-analysis}
To understand where agents fail, we classify each checkpoint into one of four capability categories: Data Retrieval, Clinical Reasoning, Action Execution, and Documentation, based on the evaluation checkpoint types.
Table~\ref{tab:failure_breakdown} reports the distribution of failed checkpoints across these categories.

\mypara{Failure distribution across capability categories.}
Clinical reasoning accounts for the majority of failures across nearly all models, confirming that multi-step clinical interpretation, rather than tool-use mechanics or data access, is the primary bottleneck. Weaker models show even higher concentration of reasoning failures than stronger ones, and GPT-5.5 reaches the lowest CR share, indicating that stronger models resolve more reasoning checkpoints and shift their failure distribution toward other categories.
Action execution failures remain stable across all models, reflecting a systematic challenge in translating clinical decisions into structured EHR writes that is largely independent of model capability.
Documentation failures are more pronounced in stronger models, likely because stronger agents attempt more complete clinical notes and are penalized for omissions that weaker models avoid by producing minimal output. Data retrieval failures are slightly elevated in weaker models that struggle to navigate the EHR environment.

% \begin{figure}[!th]
%     \centering
%     \includegraphics[width=\linewidth]{img/error_analysis_v2.pdf}
%     \caption{Fine-grained root cause analysis of Claude Opus~4.6 failures. Each failed checkpoint is classified into one of four categories and a fine-grained subtype.}
%     \label{fig:error_case_study}
% \end{figure}

% \mypara{Fine-grained root causes.}
% Figure~\ref{fig:error_case_study} drills into Claude Opus~4.6 failures with fine-grained subtypes, revealing distinct dominant patterns within each category.
% Clinical reasoning failures are largely \emph{incomplete reasoning} (41\%): the agent lands the headline diagnosis but skips the dose, monitoring plan, alternative differential, or safety caveat, gaps that coarser benchmarks would miss but our fine-grained rubric correctly flags.
% Action execution failures are dominated by an \emph{output gap} (73\%), in which the agent writes the correct action in its clinical note but never creates the corresponding FHIR resource.
% Documentation failures are largely \emph{cascade} (83\%), inheriting an upstream reasoning, action, or retrieval error rather than reflecting an independent documentation problem.
% Data retrieval failures are concentrated in \emph{not retrieved} (80\%): the required FHIR query was never issued.

\begin{figure}[!h]
    \centering
    \includegraphics[width=\linewidth]{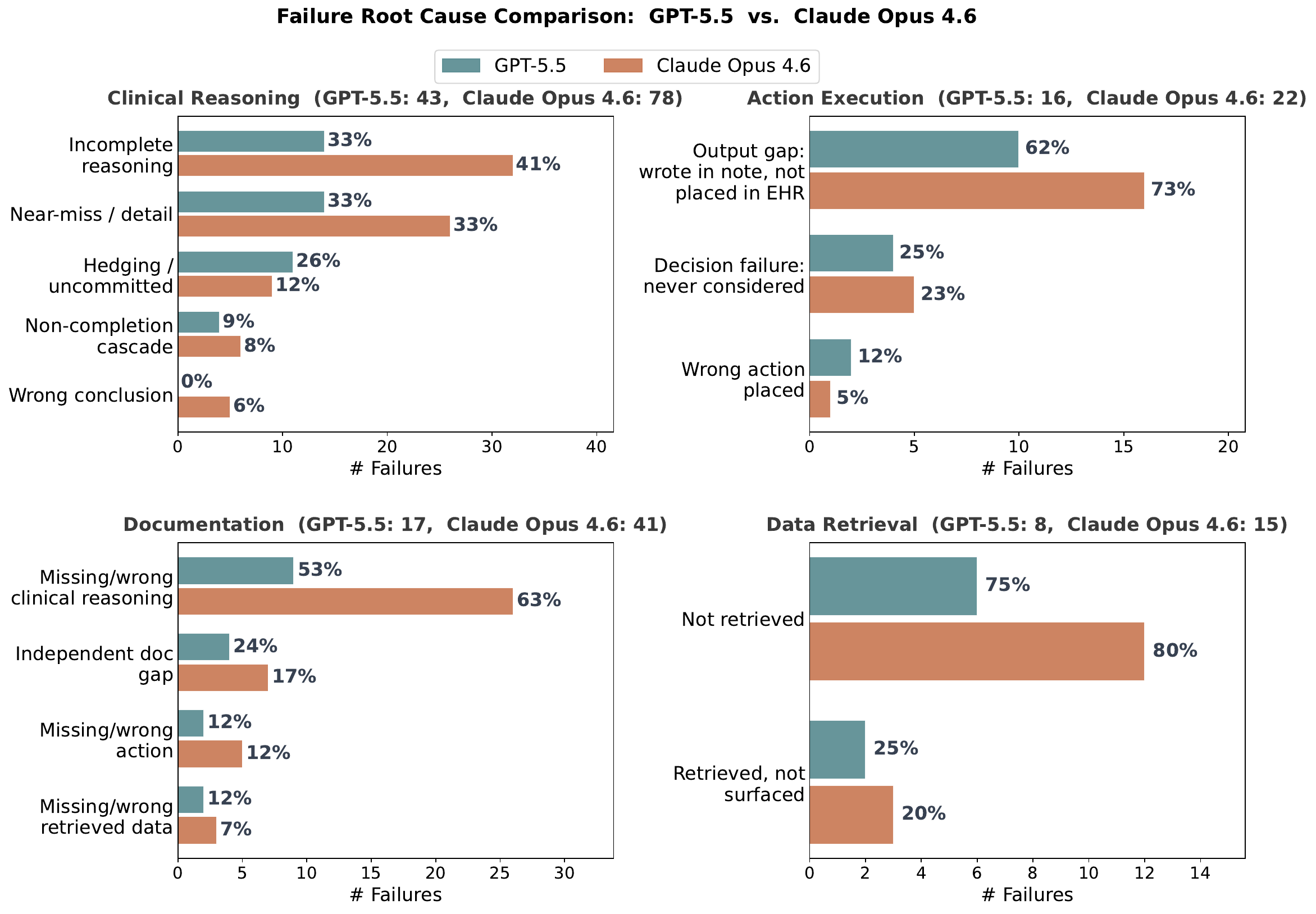}
    \caption{Fine-grained root cause comparison between GPT-5.5 and Claude Opus~4.6. Each failed checkpoint is classified into one of four categories and a fine-grained subtype; bars show the absolute count per subtype with the percentage-of-category label, allowing direct subgroup-level comparison between the two models.}
    \label{fig:error_case_study}
\end{figure}

\mypara{Fine-grained root causes.}
Figure~\ref{fig:error_case_study} contrasts fine-grained failure subtypes of GPT-5.5 and Claude Opus~4.6. Clinical Reasoning failures in both models concentrate in \emph{incomplete reasoning} and \emph{near-miss / detail}: the agent reaches the right clinical territory but drops a specific rubric element such as a dose, monitoring plan, or safety caveat. Outright \emph{wrong conclusions} are rare in both models, so when either agent commits it is rarely wrong; it just leaves rubric elements uncovered. Action Execution is dominated by an \emph{output gap} (the agent writes the correct action in its note but never creates the corresponding FHIR resource), Documentation by \emph{cascade} from upstream errors, and Data Retrieval by \emph{not retrieved} queries. The quantitative shift is uniform: GPT-5.5 records lower absolute counts in every subgroup of every category, suggesting its improvement is broad behavioral consistency rather than relief in any one failure mode.

\subsection{What Drives the GPT-5.5 Lead?}
\label{sec:gpt55-advantage}

To understand why GPT-5.5 outperforms the next-strongest model (Claude Opus~4.6) by such a wide margin, we conducted a head-to-head comparison on two models and observe that (1) EHR data fidelity: Opus~4.6 stated chart-contradictory facts (e.g., describing a patient as not on a PPI when pantoprazole is documented) or abstracted away a required numeric value, while GPT-5.5 instead surfaced the literal data; (2) GPT-5.5 reliably enumerated all required rubric items (a workup-lab set, a documentation checklist) where Opus~4.6 typically covered fewer.

The advantage is not a clinical-knowledge gap. The 8 reverse cases involve the same kinds of failures on different tasks, and the two models share a comparable wrong-conclusion rate. Mechanistically, GPT-5.5 issues markedly more tool calls on the advantage tasks, suggesting deeper EHR exploration translates into stronger data grounding. Both models thus exhibit comparable underlying clinical reasoning capability; GPT-5.5 simply applies it more consistently.

%% file: tables/results_main.tex
\begin{table*}[!t]
\caption{Main results on \bench. Pass@1 is the fraction of tasks fully completed (\%). \#Turns is the average number of tool calls per task. Bold indicates the best score in each column.}
\label{tab:main_results}
\begin{center}
\setlength{\tabcolsep}{5pt}
\adjustbox{max width=\linewidth}{
\begin{tabular}{lcccc}
\toprule
Model & Pass@1 & Pass@3 & Pass\textasciicircum{}3 & \#Turns \\
\midrule
\multicolumn{5}{c}{\textit{Proprietary Models}} \\
GPT-5.5            & \textbf{46.3} $\pm$ 1.2 & \textbf{57.4} & \textbf{28.0} & 41.9 \\
Claude Opus 4.6    & 31.7 $\pm$ 2.3          & 41.5          & 18.0          & 25.2 \\
Claude Opus 4.7    & 29.3 $\pm$ 2.5          & 37.9          & 18.0          & 16.2 \\
GPT-5.4            & 27.7 $\pm$ 1.5          & 37.7          & 13.0          & 39.8 \\
Claude Sonnet 4.6  & 23.0 $\pm$ 2.6          & 33.2          & 9.0           & 22.3 \\
MiMo-v2.5-Pro      & 16.7 $\pm$ 4.0          & 23.6          & 6.0           & 29.5 \\
Qwen3.6-Plus       & 13.7 $\pm$ 4.0          & 22.6          & 2.0           & 28.0 \\
MiniMax M2.7       & 8.7 $\pm$ 1.2           & 15.9          & 1.0           & 29.7 \\
Gemini Pro 3.1     & 6.0 $\pm$ 1.0           & 9.3           & 3.0           & 30.4 \\
Grok-4.20          & 5.3 $\pm$ 3.2           & 9.7           & 1.0           & 16.7 \\
\midrule
\multicolumn{5}{c}{\textit{Open-Source Models}} \\
DeepSeek V4-Pro    & 18.7 $\pm$ 2.9          & 27.9          & 6.0           & 35.3 \\
Kimi-K2.6          & 17.0 $\pm$ 2.6          & 26.3          & 5.0           & 42.4 \\
\bottomrule
\end{tabular}
}
\end{center}
\end{table*}

%% file: tables/results_speciality_breakdown.tex
\begin{table*}[!t]
\caption{Pass@1 (\%) by clinical specialty group. Bold indicates the best score in each column.}
\label{tab:specialty_breakdown}
\begin{center}
\setlength{\tabcolsep}{3pt}
\adjustbox{max width=\linewidth}{
\begin{tabular}{lccccccccc}
\toprule
Model & Cardio & Endo & GI/Hep & Immunol/ID & Psych/Neuro & Hem/Onc & Neph/Urol & Pulm/Other & \textbf{Overall} \\
\midrule
\multicolumn{10}{c}{\textit{Proprietary Models}} \\
GPT-5.5            & \textbf{55.6} & \textbf{59.0} & \textbf{57.1} & \textbf{38.9} & \textbf{33.3} & \textbf{48.7} & 29.2          & \textbf{48.1} & \textbf{46.3} \\
Claude Opus 4.6    & 27.8          & 35.9          & 35.7          & \textbf{38.9} & 27.1          & 30.8          & \textbf{33.3} & 25.9          & 31.7          \\
Claude Opus 4.7    & 38.9          & 28.2          & 28.6          & 22.2          & 18.8          & 30.8          & \textbf{33.3} & 38.9          & 29.3          \\
GPT-5.4            & 27.8          & 30.8          & 21.4          & 27.8          & 22.9          & 38.5          & 20.8          & 29.6          & 27.7          \\
Claude Sonnet 4.6  & 33.3          & 10.3          & 26.2          & 27.8          & 25.0          & 25.6          & \textbf{33.3} & 14.8          & 23.0          \\
MiMo-v2.5-Pro      & 11.1          & 10.3          & 23.8          & 16.7          & 27.1          & 17.9          & 4.2           & 13.0          & 16.7          \\
Qwen3.6-Plus       & 5.6           & 12.8          & 9.5           & 16.7          & 20.8          & 10.3          & 12.5          & 14.8          & 13.7          \\
MiniMax M2.7       & 0.0           & 5.1           & 11.9          & 11.1          & 8.3           & 7.7           & 4.2           & 13.0          & 8.7           \\
Gemini Pro 3.1     & 5.6           & 10.3          & 7.1           & 0.0           & 8.3           & 5.1           & 0.0           & 7.4           & 6.0           \\
Grok-4.20          & 5.6           & 5.1           & 9.5           & 2.8           & 10.4          & 2.6           & 4.2           & 1.9           & 5.3           \\
\midrule
\multicolumn{10}{c}{\textit{Open-Source Models}} \\
DeepSeek V4-Pro    & 16.7          & 7.7           & 33.3          & 19.4          & 14.6          & 15.4          & 12.5          & 24.1          & 18.7          \\
Kimi-K2.6          & 27.8          & 12.8          & 21.4          & 16.7          & 14.6          & 17.9          & 12.5          & 16.7          & 17.0          \\
\bottomrule
\end{tabular}
}
\end{center}
\end{table*}

%% file: tables/results_failure_distribution.tex
\begin{table*}[!t]
\caption{Distribution of failed checkpoints by failure category (\%). Each failed checkpoint is classified by its evaluation type.}
\label{tab:failure_breakdown}
\begin{center}
\setlength{\tabcolsep}{5pt}
\adjustbox{max width=\linewidth}{
\begin{tabular}{lcccc}
\toprule
 & \multicolumn{4}{c}{\textbf{Failure Distribution (\% of Failed Checkpoints)}} \\
\cmidrule(lr){2-5}
Model & Data Retrieval & Clinical Reasoning & Action Execution & Documentation \\
\midrule
\multicolumn{5}{c}{\textit{Proprietary Models}} \\
GPT-5.5              & 10.0 & 42.5 & 19.4 & 28.1 \\
Claude Opus 4.6      & 10.0 & 50.4 & 17.4 & 22.2 \\
Claude Opus 4.7      & 14.2 & 48.3 & 14.3 & 23.2 \\
GPT-5.4              & 10.3 & 46.1 & 20.6 & 23.0 \\
Claude Sonnet 4.6    & 11.8 & 50.2 & 18.8 & 19.2 \\
MiMo-v2.5-Pro        & 12.6 & 52.2 & 17.5 & 17.7 \\
Qwen3.6-Plus         & 11.6 & 53.8 & 19.3 & 15.3 \\
MiniMax M2.7         & 13.7 & 50.4 & 20.0 & 15.9 \\
Gemini Pro 3.1       & 15.2 & 50.1 & 19.7 & 15.0 \\
Grok-4.20            & 13.6 & 54.4 & 18.7 & 13.4 \\
\midrule
\multicolumn{5}{c}{\textit{Open-Source Models}} \\
DeepSeek V4-Pro      & 12.4 & 58.3 & 15.3 & 14.0 \\
Kimi-K2.6            & 11.9 & 50.3 & 19.1 & 18.7 \\
\midrule
\textit{Average}     & 12.4 & 50.4 & 18.8 & 18.4 \\
\bottomrule
\end{tabular}
}
\end{center}
\end{table*}

%% file: 4_discussion.tex
\section{Discussion}
\label{sec:discussion}

We introduced \bench, a benchmark for evaluating LLM agents on real-world clinical tasks in real EHR environments.
\bench provides 100 long-horizon tasks sourced from real consultation cases, both authored and independently reviewed by physicians, with 670 fine-grained checkpoints supporting verifiable evaluation.
A comprehensive evaluation of 12 LLM agents reveals a substantial gap, with the best agents completing only 46\% of tasks and reliably solving 28\% across multiple attempts, indicating that current agents remain far from autonomously and reliably handling real-world clinical tasks.

\mypara{Limitations and future work.}
\bench currently targets EHR-mediated, e-consult-style physician work, leaving several directions for future iterations:
(i) broader health workflow coverage, extending the infrastructure to discharge summaries, medication management and other inpatient settings;
(ii) from autonomy to collaboration, evaluating agent-user interaction with simulated physicians or patients under clinical constraints and shared decision-making; and
(iii) multimodal inputs and expanded tool ecosystems, incorporating imaging, waveforms, and external knowledge bases (e.g., PubMed, RxNorm) to bring evaluation closer to the full information context of practicing physicians.
We hope \bench serves as a rigorous yardstick for tracking progress toward LLM agents that can reliably support physicians in real healthcare settings.

%% file: 7_acknowledgement.tex
\section*{Acknowledgments}
R.L. is supported by the Stanford Center for Asian Health Research and Education (CARE) Postdoctoral Scholarship (\url{https://asianhealth.stanford.edu/}).

%% file: 9_appendix.tex
\clearpage

\appendix
\renewcommand{\thefigure}{A\arabic{figure}}
\setcounter{figure}{0}
\renewcommand{\thetable}{A\arabic{table}}
\setcounter{table}{0}
\setcounter{equation}{0}

\section*{Appendix}

\section{Task Taxonomy}
\label{app:taxonomy}

\bench comprises 100 clinician-validated tasks spanning 21 subspecialties organized into 8 specialty groups, and 4 task types subdivided into 14 fine-grained subtypes.
Specialty groups reflect the primary clinical domain of each task, and task types follow a reformulation of the \textsc{DICTUM} clinical decision-making taxonomy~\citep{ofstad2016medical}: \emph{Diagnosis \& Interpretation} (result and imaging interpretation), \emph{Medication Prescribing} (new therapy, regimen adjustment, acute treatment), \emph{Treatment Planning} (longer-horizon planning including medication optimization, empiric treatment, diagnostic planning, substance-use management, and risk--benefit analysis), and \emph{Workup \& Risk Stratification} (disease workup, abnormal lab evaluation, incidental-finding follow-up, infection evaluation).

We provide two complementary enumerations of the full task inventory:
Table~\ref{tab:task_taxonomy_by_type} groups tasks by task type and fine-grained subtype;
Table~\ref{tab:task_taxonomy_by_specialty} expands to all 21 subspecialties.
Figure~\ref{fig:task_distribution} (main paper) summarizes the marginal distributions over specialty and task type.

\input{tables/task_taxonnomy_by_type}

\input{tables/task_taxonomy_by_specialty}

\section{Checkpoint Grader Specifications}
\label{app:graders}

Each checkpoint in \bench is graded by one of three grader types, selected based on the nature of the criterion being assessed. We provide full specifications below.

\paragraph{Code graders}
verify agent actions by querying the FHIR server for resources created during execution. The validation logic checks resource type, status, coded values, and dosage parameters. For example, a medication order checkpoint asserts that a \texttt{MedicationRequest} exists matching the expected drug name pattern, dose range, and frequency. Date filtering ensures that only agent-created resources (those authored after the task timestamp) are evaluated, distinguishing agent actions from pre-existing patient data.

\paragraph{Hybrid graders}
combine programmatic verification with LLM-assisted extraction for checkpoints where the ground truth is objectively defined but the agent's output format varies. First, ground-truth values are computed deterministically from FHIR data (e.g., calculating a CHA$_2$DS$_2$-VASc score from patient demographics and coded diagnoses). Then, an LLM extracts the corresponding value from the agent's free-text output. Finally, the extracted value is compared against the ground truth using deterministic logic. This approach accommodates variability in how agents report numerical results or clinical decisions while maintaining objective scoring criteria.

\paragraph{LLM-judge graders}
evaluate aspects of clinical reasoning and documentation that resist purely programmatic assessment. Each LLM-judge checkpoint specifies a detailed rubric containing: (i)~required clinical elements with ground-truth values, (ii)~evaluation criteria organized as sequential verification steps, and (iii)~three-level scoring thresholds (\textsc{pass} / \textsc{partial} / \textsc{fail}). For instance, a documentation checkpoint rubric may require that the agent's note contains correct lab values, cites appropriate clinical guidelines, provides patient-specific rationale for treatment decisions, and avoids clinically unsafe recommendations.

\section{Data Annotation Protocol}
\label{app:annotation}

\begin{figure}[!ht]
    \centering
    \includegraphics[width=\linewidth]{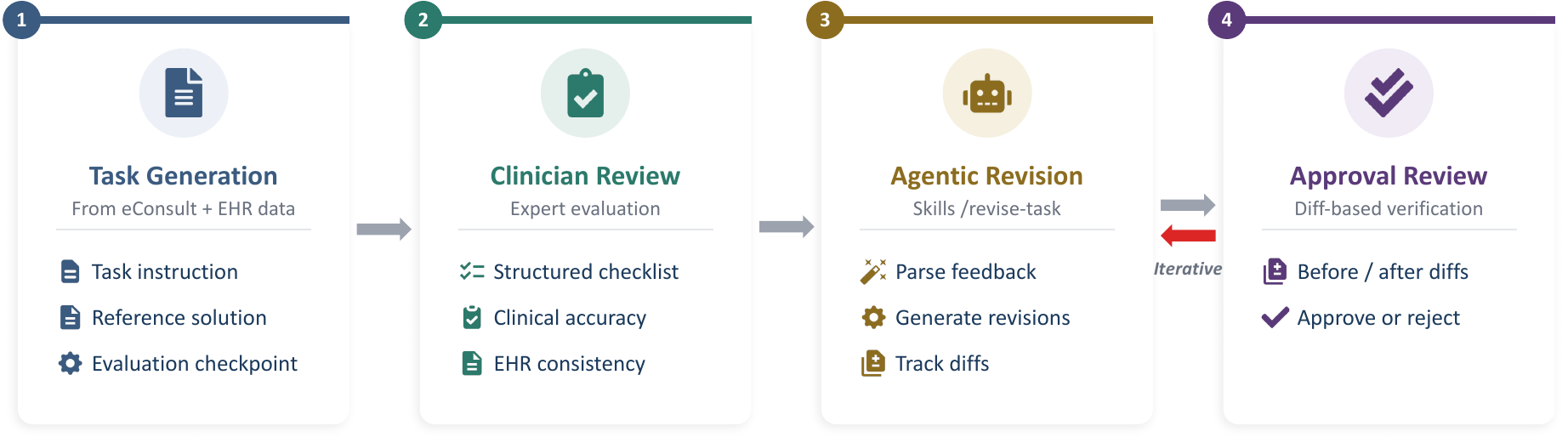}
    \caption{Clinician validation pipeline. Each task progresses through four stages: (1)~LLM-assisted generation of draft artifacts, (2)~initial expert review against a structured checklist, (3)~agentic revision that produces targeted edits with rationale, and (4)~approval review where the clinician accepts or rejects each change. Stages~(3)--(4) iterate until convergence.}
    \label{fig:annotation_pipeline}
\end{figure}

\subsection{Data Collection}

Source clinical scenarios are drawn from de-identified electronic health records (EHR) in the STARR~\citep{datta2020new}, specifically from the e-consult system in which primary care providers submit structured referral questions to subspecialists. E-consult records were selected to span diverse subspecialties and task types (diagnostic workup, treatment planning, medication management, referral coordination, and clinical documentation), yielding an initial pool of 185 candidate cases. Cases were filtered to ensure sufficient clinical complexity: each selected case must involve at least two distinct reasoning steps and require interaction with multiple EHR data sources (e.g., labs, medications, clinical notes).

\subsection{Patient Data De-identification}
\label{app:deidentification}

\paragraph{Structured EHR data.}
Structured fields (laboratory values, vital signs, imaging measurements) were perturbed while preserving clinical plausibility: continuous values were shifted by a random noise drawn from a uniform distribution bounded by the assay's reference range, ensuring the perturbed value remains within normal or clinically expected bounds.
Categorical fields carrying task-relevant information (diagnoses, active medications, procedure codes) were further anonymized by randomly dropping a subset of entries and inserting clinically plausible but task-irrelevant values drawn from the same ontology, preventing the agent from trivially identifying the target condition from the record structure alone.
Patient demographics (name, date of birth, address, phone number, MRN) were replaced with synthetically generated values using the Faker library\footnote{\url{https://pypi.org/project/Faker/}}, producing internally consistent but entirely fictitious identities.

\paragraph{Unstructured clinical notes.}
Free-text clinical notes were processed using Stanford SecureGPT, a series of PHI-safe LLMs operating within a secure institutional enclave that does not transmit data externally.
SecureGPT was prompted to (i) identify and remove all protected health information (PHI) categories defined under HIPAA Safe Harbor, and (ii) rewrite the note in a stylistically distinct form that preserves the clinical meaning and key findings while producing text that differs substantially from the original. 

\subsection{Annotation Guidelines and Quality Control}
\label{app:annotation-guidelines}

Each task artifact, comprising a natural-language instruction, a solution summary, structured evaluation checkpoints, and an EHR reference, was reviewed by a domain-expert clinician through a dedicated web-based annotation application.
Reviewers were provided with a structured checklist covering five dimensions:

\begin{enumerate}
    \item \textbf{Instruction clarity.} Whether the task instruction is unambiguous and actionable for an autonomous agent, with a single correct interpretation. Reviewers flagged ambiguities and suggested specific rewording.

    \item \textbf{Clinical reasoning validity.} Whether the reasoning chain in the solution summary is medically correct and follows accepted clinical guidelines. Reviewers flagged flawed or incomplete reasoning and provided the correct clinical logic.

    \item \textbf{Completeness.} Whether the solution omits clinically important elements that a competent clinician would address (e.g., safety checks, guideline-mandated follow-up, contraindication screening).

    \item \textbf{Patient safety.} Whether any recommended action could cause patient harm. Tasks with safety concerns were flagged and revised or excluded.

    \item \textbf{EHR consistency.} Whether all clinical values, dates, and findings cited in the solution match the de-identified patient record. Reviewers verified specific data points against the EHR reference panel.
\end{enumerate}

Reviewers selected a categorical response for each checklist item (e.g., \emph{Clear / Ambiguous / Missing context}) and were required to provide free-text justification for any non-positive response, with access to the full de-identified EHR in Excel format and a curated summary panel. Each review session took approximately 20 minutes per task. Quality was enforced through the iterative human-in-the-loop revision pipeline described in \S\ref{sec:validation}; on average, tasks required at least two review rounds before finalization.

\begin{figure}[!ht]
    \centering
    \includegraphics[width=\linewidth]{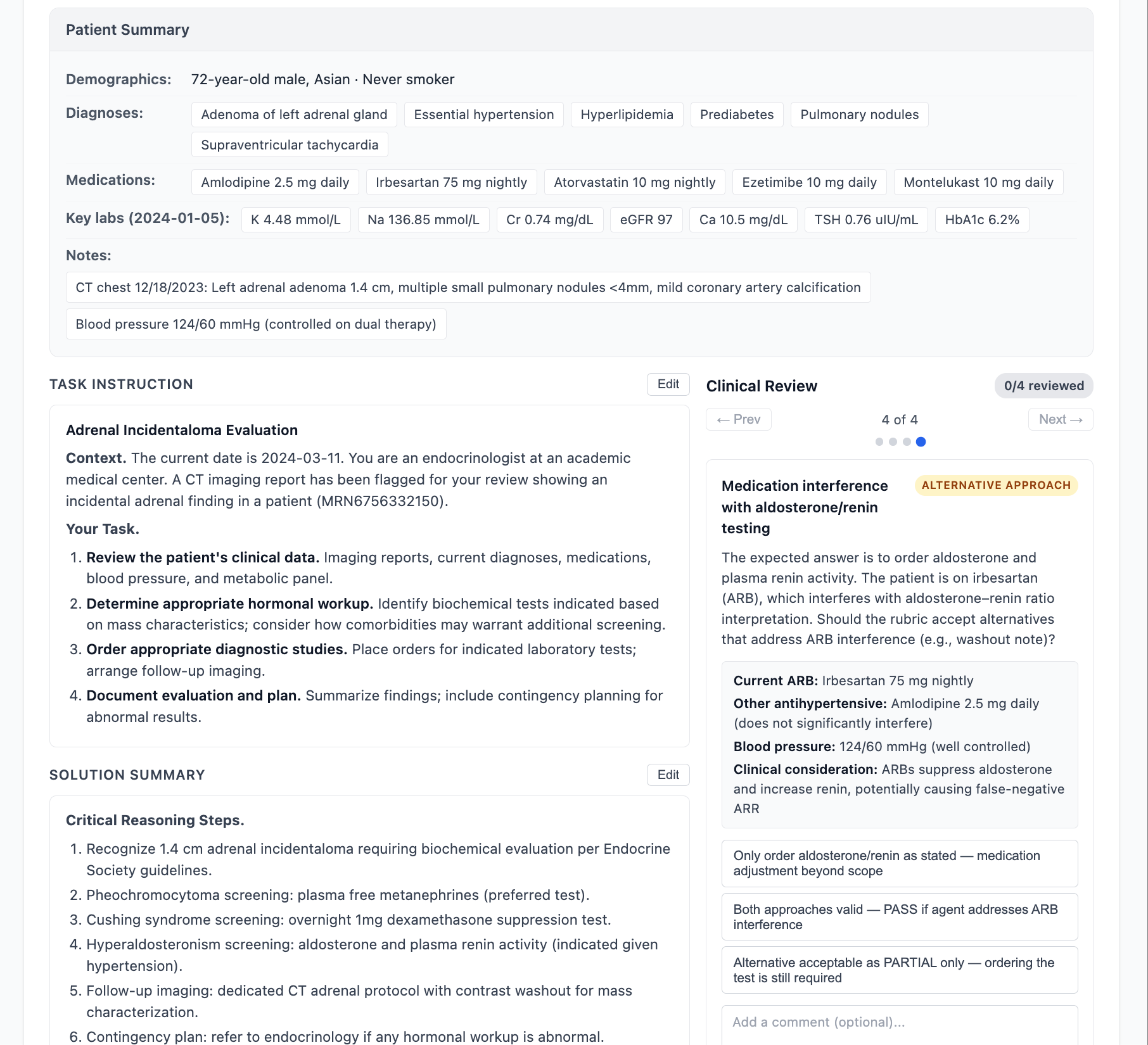}
    \caption{Initial-review screen (stage~2) of the annotation web application: structured clinical-review checklist (right) alongside the task instruction, solution summary, and patient summary panel (left and top).}
    \label{fig:annotation_ui_initial}
\end{figure}

\begin{figure}[!ht]
    \centering
    \includegraphics[width=\linewidth]{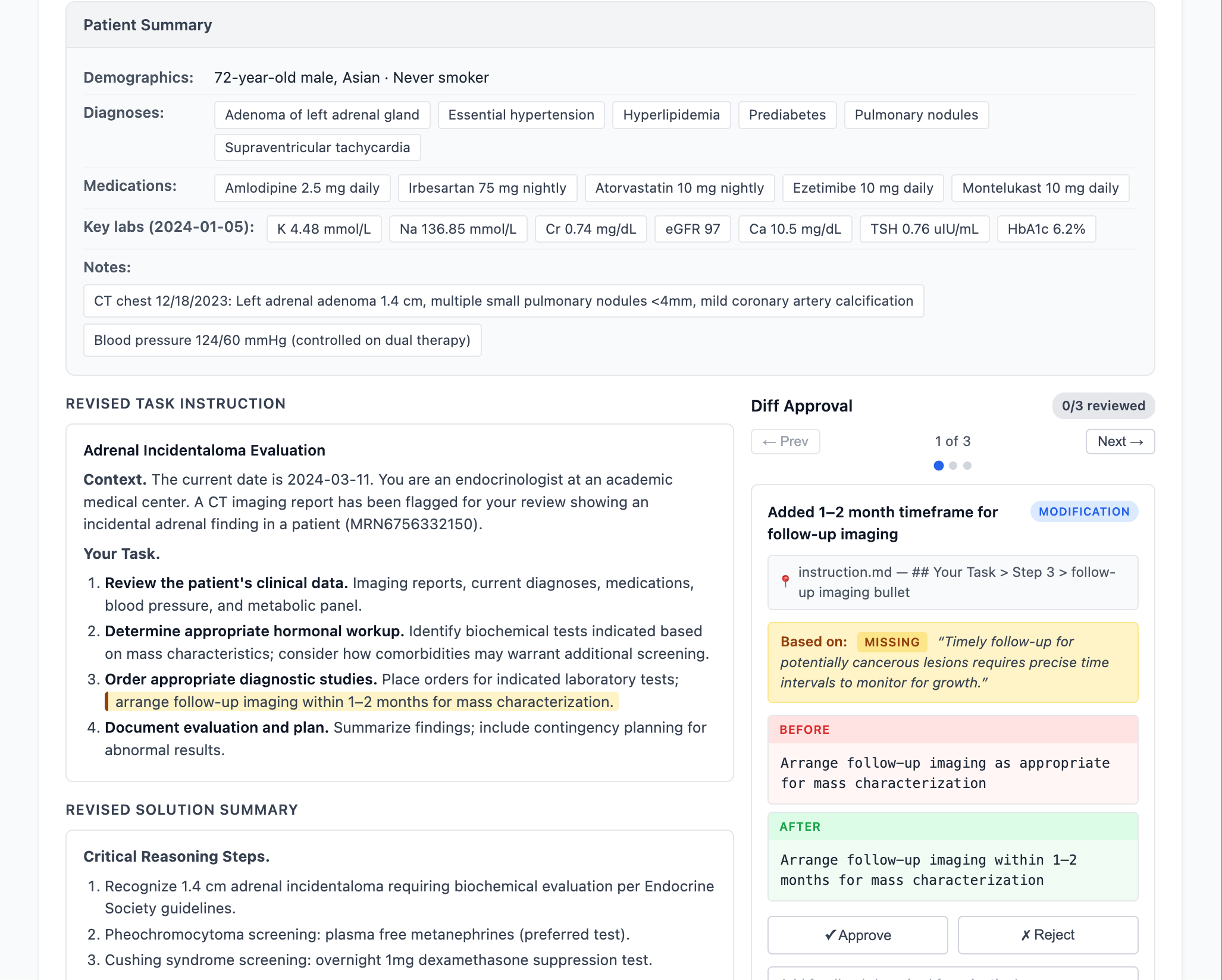}
    \caption{Diff-approval screen (stage~4) of the annotation web application: agent-proposed edits are shown as before/after diffs with the originating reviewer comment (right), and individually approved or rejected by the clinician.}
    \label{fig:annotation_ui_diff}
\end{figure}

Figures~\ref{fig:annotation_ui_initial}~and~\ref{fig:annotation_ui_diff} show the two main interactive screens of the annotation web application. Figure~\ref{fig:annotation_ui_initial} is the initial-review screen used in stage~(2), where the clinician inspects the task instruction, solution summary, and a curated patient summary alongside a structured clinical-review checklist. Figure~\ref{fig:annotation_ui_diff} is the diff-approval screen used in stage~(4), where each agent-proposed edit is presented as a before/after diff with the originating reviewer comment, and the clinician approves or rejects it.

\section{Annotator Details}
\label{app:annotators}

The benchmark was validated by 11 clinicians recruited through institutional networks at Stanford. Because every task is derived from an e-consult case whose clinical answer was already authored by a board-certified subspecialist, our annotators function as second-tier reviewers rather than de novo authors of subspecialty judgments. Their task is to verify that the generated instruction, solution summary, and checkpoints faithfully reflect the subspecialist's original answer, that the reasoning remains clinically coherent, and that no patient-safety issue was introduced during task construction. Table~\ref{tab:annotators} summarizes the cohort's demographics.

\begin{table}[!ht]
\caption{Demographic summary of the 11 expert annotators.}
\label{tab:annotators}
\centering
\setlength{\tabcolsep}{5pt}
\renewcommand{\arraystretch}{1.15}
\begin{tabular}{llc}
\toprule
\textbf{ID} & \textbf{Department} & \textbf{Years of Experience} \\
\midrule
A1  & Internal Medicine / Hospital Medicine         & 12 \\
A2  & Internal Medicine                             & 10 \\
A3  & Internal Medicine                             &  8 \\
A4  & Internal Medicine                             &  7 \\
A5  & Internal Medicine                             &  6 \\
A6  & Internal Medicine                             &  5 \\
A7  & Internal Medicine                             &  5 \\
A8  & Internal Medicine                             &  6 \\
A9  & Cardiology / Internal Medicine                &  6 \\
A10 & Emergency Medicine / Clinical Informatics     &  5 \\
A11 & Psychiatry                                    & 
5
\\
\bottomrule
\end{tabular}
\end{table}

\section{Agent Framework Implementation}
\label{app:agent-framework}

The agent framework described in \S\ref{sec:environment} is implemented as a $\sim$300-line Python module wrapping the OpenAI-compatible chat completions API. We intentionally avoid third-party agent SDKs (e.g., OpenAI Agents SDK, LangChain, Claude Code) so that performance differences between models are attributable to model capability rather than model-specific scaffolding. Below we describe the interaction loop, robustness enhancements, and configuration settings used throughout our experiments.

\paragraph{Interaction loop.}
At each step $t$, the agent state consists of the system prompt, the task instruction, and the running message history $(m_1, \dots, m_{t-1})$. The framework invokes the model with (i)~the message history, (ii)~the JSON schemas of all 14 tools defined in \S\ref{sec:environment}, and (iii)~the provider-specific reasoning effort (\texttt{high} for reasoning-capable models; unset otherwise). Temperature is left at the provider's default. If the response contains no tool calls, the text content is treated as the agent's final answer and the loop terminates. Otherwise, each tool call is dispatched to the corresponding handler, the resulting FHIR JSON bundle or file-write confirmation is serialized and appended as a tool-role message, and the loop repeats. Parallel tool calls within a single step are supported natively and executed sequentially by the handler.

\paragraph{Tool inventory.}
Table~\ref{tab:tool_inventory} enumerates the complete tool interface exposed to agents. The action space contains 14 registered tools: 13 FHIR-backed EHR tools and one workspace file-writing tool.

\begin{table}[!htbp]
\caption{Tool inventory exposed to agents, named by the underlying FHIR resource. Search tools issue \texttt{GET} requests against the FHIR server; create tools issue \texttt{POST} requests that persist on the task-local server.}
\label{tab:tool_inventory}
\centering
\small
\setlength{\tabcolsep}{6pt}
\renewcommand{\arraystretch}{1.2}
\begin{tabular}{p{0.34\linewidth}p{0.58\linewidth}}
\toprule
\textbf{Resource / endpoint} & \textbf{Key arguments} \\
\midrule
\multicolumn{2}{l}{Search (\texttt{GET})} \\
\midrule
\texttt{Patient}                       & \texttt{identifier}, \texttt{name}, \texttt{birthdate}, \texttt{gender}, \texttt{telecom} \\
\texttt{Condition}                     & \texttt{patient}, \texttt{clinical\_status}, \texttt{category}, \texttt{code}, \texttt{onset\_date} \\
\texttt{Observation} (labs)            & \texttt{patient}, \texttt{category}, \texttt{code}, \texttt{date} \\
\texttt{Observation} (vitals)          & \texttt{patient}, \texttt{code}, \texttt{category}, \texttt{date} \\
\texttt{Observation} (social history)  & \texttt{patient}, \texttt{code}, \texttt{date} \\
\texttt{MedicationRequest}             & \texttt{patient}, \texttt{status}, \texttt{intent}, \texttt{authoredon}, \texttt{date} \\
\texttt{Procedure}                     & \texttt{patient}, \texttt{category}, \texttt{date} \\
\texttt{DocumentReference}             & \texttt{patient}, \texttt{encounter}, \texttt{date}, \texttt{category}, \texttt{type} \\
\texttt{ServiceRequest}                & \texttt{patient}, \texttt{status}, \texttt{intent}, \texttt{category}, \texttt{code}, \texttt{authored} \\
\midrule
\multicolumn{2}{l}{Create (\texttt{POST})} \\
\midrule
\texttt{MedicationRequest}             & \texttt{patient}, \texttt{medication\_display}, \texttt{dose\_value}, \texttt{frequency\_text} \\
\texttt{ServiceRequest}                & \texttt{patient}, \texttt{code\_code}, \texttt{code\_display}, \texttt{priority}, \texttt{reason\_display} \\
\texttt{Appointment}                   & \texttt{patient}, \texttt{practitioner}, \texttt{start}, \texttt{end}, \texttt{based\_on} \\
\texttt{Communication}                 & \texttt{patient}, \texttt{message\_text}, \texttt{recipient}, \texttt{sender} \\
\midrule
\multicolumn{2}{l}{Utility} \\
\midrule
\texttt{write\_file}                   & \texttt{file\_path}, \texttt{content}, \texttt{mode} \\
\bottomrule
\end{tabular}
\end{table}

\paragraph{Robustness enhancements.}
We add four safeguards beyond the basic loop:

\begin{itemize}[leftmargin=*,nosep]
    \item \textbf{Tool error handling.} Exceptions raised during tool dispatch (e.g., FHIR server unreachable, malformed \texttt{resourceType}), or JSON decoding errors on tool arguments, are caught and returned to the model as a structured \texttt{\{"error": ...\}} message rather than propagating. This allows the agent to observe and recover from transient or semantic errors.
    \item \textbf{Overlong-output handling.} Tool responses exceeding \texttt{MAX\_TOOL\_OUTPUT\_LEN}=10{,}000 characters are truncated with an explicit tail hint: ``output truncated, showing first $N$ of $M$ characters; use filters such as `code', `date=ge...', or reduced `count' to narrow results.'' This prevents single large FHIR bundles (e.g., unfiltered \texttt{Observation} search) from consuming the entire context window while instructing the agent how to recover.
    \item \textbf{Loop detection.} The agent is aborted early under any of three stuck-state conditions, each with a five-step threshold: (a)~\texttt{MAX\_REPEATED\_ERRORS}, the same tool returns the same error for five consecutive calls; (b)~\texttt{MAX\_REPEATED\_CALLS}, the same \texttt{(tool, args, output)} triple is observed five times in a row; (c)~\texttt{MAX\_REPEATED\_BATCHES}, the same multi-call batch (as an unordered set) repeats five times within a sliding window of ten steps. An additional novelty counter aborts when no new \texttt{(tool, args)} pair has been issued for 15 consecutive steps. 
    \item \textbf{Reasoning-effort pass-through.} For models supporting the \texttt{reasoning\_effort} parameter, we pass \texttt{high} to maximize reasoning budget. For models without native reasoning support, the parameter is omitted.
\end{itemize}

% \paragraph{Trajectory logging.}
% Every interaction is recorded to a JSONL trajectory file containing, per event: the event type (\texttt{llm\_response}, \texttt{tool\_call}, \texttt{final\_result}, or \texttt{error}), the full model message (including reasoning traces when available via \texttt{reasoning} or \texttt{reasoning\_details} fields), the complete tool input and output, token counts, and finish reason. These logs drive both the checkpoint evaluation (\S\ref{sec:evaluation}) and the error analysis (\S\ref{sec:error-analysis}).

\paragraph{System prompt.}
The system prompt is intentionally minimal to avoid steering model behavior. Task-specific details (the working directory for file outputs and the FHIR server URL substitutions) are appended to the per-task instruction rather than hard-coded in the system prompt, keeping the latter reusable across tasks:

\begin{tcolorbox}[
    enhanced,
    colback=gray!5,
    colframe=gray!60,
    arc=1mm,
    left=2mm, right=2mm, top=1.5mm, bottom=1.5mm,
    fontupper=\small\ttfamily,
    breakable
]
You are a clinical AI assistant designed to support healthcare professionals.\\
You have access to an EHR system via FHIR API tools and can write files to disk.

\vspace{0.5em}
Guidelines:
\begin{itemize}[nosep,leftmargin=*,label=-]
    \item Use the FHIR search tools to retrieve patient data before making clinical decisions.
    \item Use the FHIR create tools to place orders, send messages, or schedule appointments.
    \item Use the write\_file tool to save deliverables (notes, assessments, reports) to disk.
    \item Be thorough: retrieve all relevant clinical data before writing your assessment.
    \item Be accurate: base your clinical reasoning on the actual patient data retrieved.
    \item Complete all tasks specified in the instruction before finishing.
\end{itemize}
\end{tcolorbox}

\paragraph{Configuration.}
Unless otherwise noted, all experiments use the provider's default temperature, \texttt{max\_steps=100}, \texttt{parallel\_tool\_calls=True}, tool-output truncation at 10{,}000 characters, and provider-specific reasoning effort \texttt{high}. Each task runs against a fresh FHIR server container, eliminating state leakage across tasks.

\section{Additional Results}
\label{app:results}

\input{tables/results_task_type_breakdown}

\section{Representative Failure Trajectory}
\label{app:case_study}

To illustrate how the four error categories manifest within a single agent run, we show a reproduces a representative failure trajectory from Claude Opus~4.6 on  task ``Adrenal Insufficiency Management''. The task asks an endocrinologist agent to evaluate replacement adequacy in a 69-year-old woman with secondary adrenal insufficiency, a pituitary-axis condition diagnosed from chronic ophthalmic corticosteroid exposure and opioid use, in which aldosterone secretion is preserved.

The agent successfully retrieves the relevant \texttt{Condition}, \texttt{MedicationRequest}, and \texttt{Observation} resources, but then confidently reclassifies the clinical picture as primary adrenal insufficiency (Addison's disease) based on an elevated renin level, and recommends initiating fludrocortisone, a mineralocorticoid that is not indicated in secondary AI and can cause hypertension, hypokalemia, and volume overload. It subsequently names an urgent cardiology referral in the written plan but never creates the corresponding FHIR \texttt{ServiceRequest}, and the final documentation inherits the primary-vs-secondary misdiagnosis. A single trajectory thus exercises all four top-level failure categories: \emph{Data Retrieval} (E27.40 retrieved from FHIR but not surfaced in the output), \emph{Clinical Reasoning} (wrong conclusion: primary vs.\ secondary AI), \emph{Action Execution} (output gap: referral named but not ordered), and \emph{Documentation} (cascaded misclassification).

\input{img/case_study_task106}

%% file: tables/task_taxonnomy_by_type.tex
% Auto-generated by scripts/build_taxonomy_appendix_table.py
\begin{table}[!ht]
\caption{Task taxonomy grouped by \textbf{task type} and fine-grained subtype (100 tasks total).}
\label{tab:task_taxonomy_by_type}
\begin{center}
\setlength{\tabcolsep}{6pt}
\renewcommand{\arraystretch}{1.15}
\begin{tabular}{@{}llc@{}}
\toprule
\textbf{Task Type} & \textbf{Subtype} & \textbf{N} \\
\midrule
Diagnosis \& Interpretation (N=13) & Result Interpretation & 9 \\
 & Imaging Interpretation & 4 \\
\midrule
Medication Prescribing (N=26) & New Therapy Initiation & 9 \\
 & Regimen Adjustment & 9 \\
 & Acute Treatment & 8 \\
\midrule
Treatment Planning (N=27) & Medication Optimization & 8 \\
 & Empiric Treatment & 7 \\
 & Diagnostic Planning & 5 \\
 & Substance Use Management & 4 \\
 & Risk-Benefit Analysis & 3 \\
\midrule
Workup \& Risk Stratification (N=34) & Disease Workup & 14 \\
 & Abnormal Lab Evaluation & 11 \\
 & Incidental Finding Followup & 5 \\
 & Infection Evaluation & 4 \\
\bottomrule
\end{tabular}
\end{center}
\end{table}

%% file: tables/task_taxonomy_by_specialty.tex
% Auto-generated by scripts/build_taxonomy_appendix_table.py
\begin{table}[t]
\caption{Clinical-area coverage across 21 subspecialties. Each task is assigned its primary clinical area based on the topic.}
\label{tab:task_taxonomy_by_specialty}
\begin{center}
\setlength{\tabcolsep}{6pt}
\renewcommand{\arraystretch}{1.15}
\begin{tabular}{lc}
\toprule
\textbf{Subspecialty} & \textbf{N} \\
\midrule
Endocrinology & 10 \\
Hematology & 10 \\
Family Medicine & 9 \\
Gastroenterology & 9 \\
Infectious Disease & 7 \\
Cardiology & 6 \\
Internal Medicine & 6 \\
Psychiatry & 5 \\
Gynecology & 4 \\
Hepatology & 4 \\
Primary Care & 4 \\
Pulmonology & 4 \\
Allergy \& Immunology & 3 \\
Geriatric Medicine & 3 \\
Nephrology & 3 \\
Neurology & 3 \\
Otolaryngology & 3 \\
Urology & 3 \\
Dermatology & 2 \\
Oncology & 1 \\
Rheumatology & 1 \\
\bottomrule
\end{tabular}
\end{center}
\end{table}

%% file: tables/results_task_type_breakdown.tex
\begin{table*}[t]
\caption{Pass@1 (\%) by task type, averaged over 3 runs. Bold indicates the best score in each column.}
\label{tab:task_type_breakdown}
\begin{center}
\setlength{\tabcolsep}{3pt}
\adjustbox{max width=\linewidth}{
\begin{tabular}{lccccc}
\toprule
Model & Diagnosis/Interpretation & Med Prescribe & Treatment Plan & Workup/Risk Stratification & \textbf{Overall} \\
\midrule
\multicolumn{6}{c}{\textit{Proprietary Models}} \\
GPT-5.5            & \textbf{46.2} & \textbf{41.0} & \textbf{40.7} & \textbf{54.9} & \textbf{46.3} \\
Claude Opus 4.6    & 43.6          & 28.2          & 21.0          & 38.2          & 31.7          \\
Claude Opus 4.7    & 41.0          & 26.9          & 22.2          & 32.4          & 29.3          \\
GPT-5.4            & 35.9          & 26.9          & 23.5          & 28.4          & 27.7          \\
Claude Sonnet 4.6  & 35.9          & 24.4          & 12.3          & 25.5          & 23.0          \\
MiMo-v2.5-Pro      & 20.5          & 14.1          & 14.8          & 18.6          & 16.7          \\
Qwen3.6-Plus       & 15.4          & 12.8          & 18.5          & 9.8           & 13.7          \\
MiniMax M2.7       & 7.7           & 11.5          & 6.2           & 8.8           & 8.7           \\
Gemini Pro 3.1     & 5.1           & 5.1           & 4.9           & 7.8           & 6.0           \\
Grok-4.20          & 2.6           & 7.7           & 2.5           & 6.9           & 5.3           \\
\midrule
\multicolumn{6}{c}{\textit{Open-Source Models}} \\
DeepSeek V4-Pro    & 12.8          & 14.1          & 14.8          & 27.5          & 18.7          \\
Kimi-K2.6          & 23.1          & 10.3          & 16.0          & 20.6          & 17.0          \\
\bottomrule
\end{tabular}
}
\end{center}
\end{table*}

%% file: img/case_study_task106.tex
% ============================================================================
% Case study trajectory fragment, intended to be \input{} from appendix.tex.
%
% The main paper PREAMBLE (before \begin{document}) must contain:
%   \usepackage[most]{tcolorbox}   % [most] is REQUIRED (enables skins + breakable)
%   \usepackage{xcolor}
%   \usepackage{fontawesome5}
%
% Optional (to match the standalone demo styling):
%   \usepackage{inconsolata}   \usepackage{lmodern}
%   \usepackage[T1]{fontenc}   \usepackage{microtype}
%
% Everything below (colors / box styles / macros) can safely live in the
% document body and will execute when this file is \input'd.
% ============================================================================

% ------------------------------------------------
% Palette
% ------------------------------------------------
\definecolor{pageaccent}{RGB}{108,116,128}

\definecolor{titlebg}{RGB}{108,118,130}
\definecolor{titlefg}{RGB}{248,249,250}

\definecolor{userbg}{RGB}{238,246,244}
\definecolor{userframe}{RGB}{115,164,153}
\definecolor{usericon}{RGB}{49,108,97}

\definecolor{agentbg}{RGB}{243,242,237}
\definecolor{agentframe}{RGB}{171,162,139}
\definecolor{agenticon}{RGB}{112,104,84}

\definecolor{toolbg}{RGB}{250,243,235}
\definecolor{toolframe}{RGB}{193,154,112}
\definecolor{toolicon}{RGB}{143,95,42}

\definecolor{respbg}{RGB}{243,239,246}
\definecolor{respframe}{RGB}{150,133,170}
\definecolor{respicon}{RGB}{99,76,128}

\definecolor{codebg}{RGB}{247,247,246}
\definecolor{codeframe}{RGB}{198,198,194}
\definecolor{softtext}{RGB}{70,70,70}

\definecolor{tagA}{RGB}{175,76,72}
\definecolor{tagB}{RGB}{99,76,128}
\definecolor{tagC}{RGB}{49,108,97}
\definecolor{tagD}{RGB}{143,95,42}

% Header helpers
\newcommand{\headericon}[2]{{\large\textcolor{#1}{#2}}\hspace{0.4em}}
\newcommand{\userheader}{\headericon{usericon}{\faUserMd}\textbf{Clinician Request}\hspace{0.5em}}
\newcommand{\agentheader}{\headericon{agenticon}{\faRobot}\textbf{Agent}\hspace{0.55em}}
\newcommand{\toolheader}[1]{\headericon{toolicon}{\faTools}\textbf{#1}}
\newcommand{\respheader}{\headericon{respicon}{\faArchive}\textbf{Response}\hspace{0.5em}}

% Tag helpers (colored inline labels in response innercode blocks)
\newcommand{\condtag}{\textcolor{tagA}{[COND]}\hspace{0.35em}}
\newcommand{\medtag}{\textcolor{tagB}{[MED]}\hspace{0.35em}}
\newcommand{\labtag}{\textcolor{tagC}{[LAB]}\hspace{0.35em}}
\newcommand{\notetag}{\textcolor{tagD}{[NOTE]}\hspace{0.35em}}
\newcommand{\vittag}{\textcolor{pageaccent}{[VIT]}\hspace{0.35em}}
\newcommand{\proctag}{\textcolor{tagA}{[PROC]}\hspace{0.35em}}
\newcommand{\pattag}{\textcolor{pageaccent}{[PAT]}\hspace{0.35em}}
\newcommand{\soctag}{\textcolor{pageaccent}{[SOC]}\hspace{0.35em}}
\newcommand{\writtag}{\textcolor{tagB}{[WRITE]}\hspace{0.35em}}

\begin{titlebox}
{\large\bfseries\color{titlefg}
Task: Adrenal Insufficiency Management}\\[0.5mm]
{\small\color{titlefg!85}Agent: \texttt{claude-opus-4-6}\ \ $|$\ \ max\_steps = 100\ \ $|$\ \ reasoning\_effort = high}
\end{titlebox}

% ======================================================================
% Event 0 -- instruction
% ======================================================================
\begin{userbox}
\userheader
\# Adrenal Insufficiency Management with Symptom Evaluation

\textbf{Context.} The current date and time is 2022-06-20T07:00:00+00:00.
You are an endocrinologist at an academic medical center
(Practitioner/dr-grace-mitchell). A patient (MRN4888657619) with known
adrenal insufficiency has sent a message through the patient portal
describing worsening symptoms despite current glucocorticoid replacement
therapy. The message mentions fatigue, blood pressure instability,
elevated heart rate, and decreased appetite\ldots

\textbf{Task.} (1)~Review patient history and current medications (etiology
of the adrenal insufficiency). (2)~Assess adequacy of glucocorticoid
replacement (primary vs.~secondary AI; recent stressors; 15--25\,mg/day
physiologic range). (3)~Develop management plan with dose adjustment,
labs, referrals, contingency instructions, and save to
\texttt{/workspace/output/management\_plan.txt}.
\end{userbox}

% ======================================================================
% Event 1 -- agent_initialized
% ======================================================================
\begin{agentbox}
\agentheader
\textit{MiniAgent initialized with 14 tools.}
\end{agentbox}

% ======================================================================
% Event 2 -- llm_response  (6 parallel tool calls)
% ======================================================================
\begin{agentbox}
\agentheader
I'll start by gathering all relevant clinical data for this patient. Let
me retrieve demographics, diagnoses, medications, labs, and vitals
simultaneously.
\end{agentbox}

% Event 3
\begin{toolbox}\toolheader{fhir\_patient\_search\_demographics}\vspace{1mm}
\begin{innercode}\{"identifier": "MRN4888657619"\}\end{innercode}\end{toolbox}
\begin{responsebox}\respheader\hfill\textit{\small 1 entry}\vspace{1mm}
\begin{innercode}\pattag Patient/MRN4888657619, 69F, DOB 1953-05-09\end{innercode}\end{responsebox}

% Event 4
\begin{toolbox}\toolheader{fhir\_condition\_search\_problems}\vspace{1mm}
\begin{innercode}\{"patient": "MRN4888657619", "count": 20\}\end{innercode}\end{toolbox}
\begin{responsebox}\respheader\hfill\textit{\small 20 entries}\vspace{1mm}
\begin{innercode}\condtag E27.40 Adrenal insufficiency, unspecified\\
\condtag E03.9 Hypothyroidism, unspecified\\
\condtag E06.3 Autoimmune thyroiditis\\
\condtag I10 Essential hypertension\\
\condtag H40.23 Chronic angle-closure glaucoma, R eye\\
\ldots \textit{[+15 conditions]}\end{innercode}\end{responsebox}

% Event 5
\begin{toolbox}\toolheader{fhir\_medication\_request\_search\_orders}\vspace{1mm}
\begin{innercode}\{"patient": "MRN4888657619", "status": "active"\}\end{innercode}\end{toolbox}
\begin{responsebox}\respheader\hfill\textit{\small 13 entries}\vspace{1mm}
\begin{innercode}\medtag Hydrocortisone 10 mg PO QAM\\
\medtag Hydrocortisone 5 mg PO QPM\\
\medtag \textbf{Prednisolone acetate 1\% ophthalmic 1 drop OD qid}\\
\medtag Levothyroxine 137 mcg PO daily\\
\medtag Valsartan 80 mg PO daily\\
\ldots \textit{[+8 medications]}\end{innercode}\end{responsebox}

% Event 6
\begin{toolbox}\toolheader{fhir\_observation\_search\_labs}\vspace{1mm}
\begin{innercode}\{"patient": "MRN4888657619", "category": "laboratory", "count": 50\}\end{innercode}\end{toolbox}
\begin{responsebox}\respheader\hfill\textit{\small 50 entries}\vspace{1mm}
\begin{innercode}\labtag Na 135.1, K 4.95, Cr 0.95, eGFR 62, glucose 113.76\\
\labtag \textbf{Cortisol 20.4 mcg/dL, Renin 16.0 ng/mL/h, Aldosterone 37.0 ng/dL}\\
\labtag ALT 42.63, free metanephrine <0.20, normetanephrine 0.87\\
\ldots \textit{[+47 labs]}\end{innercode}\end{responsebox}

% Event 7
\begin{toolbox}\toolheader{fhir\_observation\_search\_vitals}\vspace{1mm}
\begin{innercode}\{"patient": "MRN4888657619", "category": "vital-signs", "count": 50\}\end{innercode}\end{toolbox}
\begin{responsebox}\respheader\hfill\textit{\small 50 entries}\vspace{1mm}
\begin{innercode}\vittag BP 118/90--164/86 (labile, wide pulse pressure)\\
\vittag HR 105--127 bpm (persistent tachycardia)\\
\vittag Temp 36.7\textdegree C\end{innercode}\end{responsebox}

% Event 8
\begin{toolbox}\toolheader{fhir\_observation\_search\_social\_history}\vspace{1mm}
\begin{innercode}\{"patient": "MRN4888657619"\}\end{innercode}\end{toolbox}
\begin{responsebox}\respheader\hfill\textit{\small 16 entries}\vspace{1mm}
\begin{innercode}\soctag Never smoker, occasional alcohol, retired anesthesiologist\end{innercode}\end{responsebox}

% ======================================================================
% Event 9 -- llm_response  (6 parallel tool calls)
% ======================================================================
\begin{agentbox}
\agentheader
Now let me search for more specific data --- the remaining conditions,
cortisol levels, and more specific vital signs and lab data.
\end{agentbox}

% Event 10
\begin{toolbox}\toolheader{fhir\_condition\_search\_problems}\vspace{1mm}
\begin{innercode}\{"patient": "MRN4888657619", "count": 50, "page\_limit": 5\}\end{innercode}\end{toolbox}
\begin{responsebox}\respheader\hfill\textit{\small 47 entries}\vspace{1mm}
\begin{innercode}\condtag Full condition list --- adrenal insuff., hypothyroid, HTN,\\
glaucoma, keratoplasty status, NSAID allergy, insomnia, \ldots\end{innercode}\end{responsebox}

% Event 11
\begin{toolbox}\toolheader{fhir\_observation\_search\_labs}\vspace{1mm}
\begin{innercode}\{"patient": "MRN4888657619", "code": "2143-6"\}\hspace{0.6em}\textit{\# cortisol}\end{innercode}\end{toolbox}
\begin{responsebox}\respheader\hfill\textit{\small 1 entry}\vspace{1mm}
\begin{innercode}\labtag Cortisol, Serum: 20.4 mcg/dL (2022-06-04, on replacement)\end{innercode}\end{responsebox}

% Event 12
\begin{toolbox}\toolheader{fhir\_observation\_search\_labs}\vspace{1mm}
\begin{innercode}\{"patient": "MRN4888657619", "code": "2951-2"\}\hspace{0.6em}\textit{\# sodium}\end{innercode}\end{toolbox}
\begin{responsebox}\respheader\hfill\textit{\small 1 entry}\vspace{1mm}
\begin{innercode}\labtag Sodium: 135.1 mmol/L (low-normal)\end{innercode}\end{responsebox}

% Event 13
\begin{toolbox}\toolheader{fhir\_observation\_search\_labs}\vspace{1mm}
\begin{innercode}\{"patient": "MRN4888657619", "code": "2823-3"\}\hspace{0.6em}\textit{\# potassium}\end{innercode}\end{toolbox}
\begin{responsebox}\respheader\hfill\textit{\small 1 entry}\vspace{1mm}
\begin{innercode}\labtag Potassium: 4.95 mmol/L (high-normal)\end{innercode}\end{responsebox}

% Event 14
\begin{toolbox}\toolheader{fhir\_observation\_search\_labs}\vspace{1mm}
\begin{innercode}\{"patient": "MRN4888657619", "code": "2947-0"\}\end{innercode}\end{toolbox}
\begin{responsebox}\respheader\hfill\textit{\small 0 entries}\end{responsebox}

% Event 15
\begin{toolbox}\toolheader{fhir\_procedure\_search\_orders}\vspace{1mm}
\begin{innercode}\{"patient": "MRN4888657619"\}\end{innercode}\end{toolbox}
\begin{responsebox}\respheader\hfill\textit{\small 18 entries}\vspace{1mm}
\begin{innercode}\proctag Penetrating keratoplasty R eye (2022-05-20)\\
\proctag Suture repair R eye (2022-05-28)\\
\ldots \textit{[+16 procedures]}\end{innercode}\end{responsebox}

% ======================================================================
% Event 16 -- llm_response  (7 parallel tool calls)
% ======================================================================
\begin{agentbox}
\agentheader
Now let me get more specific data --- searching for adrenal insufficiency
diagnoses and looking for more specific labs and the complete condition
list with relevant endocrine codes.
\end{agentbox}

% Event 17
\begin{toolbox}\toolheader{fhir\_condition\_search\_problems}\vspace{1mm}
\begin{innercode}\{"patient": "MRN4888657619", "code": "E27"\}\end{innercode}\end{toolbox}
\begin{responsebox}\respheader\hfill\textit{\small 0 entries}\end{responsebox}

% Event 18
\begin{toolbox}\toolheader{fhir\_condition\_search\_problems}\vspace{1mm}
\begin{innercode}\{"patient": "MRN4888657619", "code": "E89"\}\end{innercode}\end{toolbox}
\begin{responsebox}\respheader\hfill\textit{\small 0 entries}\end{responsebox}

% Event 19
\begin{toolbox}\toolheader{fhir\_condition\_search\_problems}\vspace{1mm}
\begin{innercode}\{"patient": "MRN4888657619", "code": "I10"\}\end{innercode}\end{toolbox}
\begin{responsebox}\respheader\hfill\textit{\small 3 entries}\vspace{1mm}
\begin{innercode}\condtag I10 Essential (primary) hypertension (x3)\end{innercode}\end{responsebox}

% Event 20
\begin{toolbox}\toolheader{fhir\_observation\_search\_labs}\vspace{1mm}
\begin{innercode}\{"patient": "MRN4888657619", "code": "3016-3"\}\end{innercode}\end{toolbox}
\begin{responsebox}\respheader\hfill\textit{\small 0 entries}\end{responsebox}

% Event 21
\begin{toolbox}\toolheader{fhir\_observation\_search\_labs}\vspace{1mm}
\begin{innercode}\{"patient": "MRN4888657619", "code": "2160-0"\}\end{innercode}\end{toolbox}
\begin{responsebox}\respheader\hfill\textit{\small 1 entry}\vspace{1mm}
\begin{innercode}\labtag Creatinine: 0.95 mg/dL\end{innercode}\end{responsebox}

% Event 22
\begin{toolbox}\toolheader{fhir\_observation\_search\_vitals}\vspace{1mm}
\begin{innercode}\{"patient": "MRN4888657619", "code": "85354-9", "date": "ge2022-05-20"\}\end{innercode}\end{toolbox}
\begin{responsebox}\respheader\hfill\textit{\small 11 entries}\vspace{1mm}
\begin{innercode}\vittag BP panel: 118/90--164/86 over 2022-05-20 to -22\end{innercode}\end{responsebox}

% Event 23
\begin{toolbox}\toolheader{fhir\_observation\_search\_vitals}\vspace{1mm}
\begin{innercode}\{"patient": "MRN4888657619", "code": "8867-4", "date": "ge2022-05-20"\}\end{innercode}\end{toolbox}
\begin{responsebox}\respheader\hfill\textit{\small 11 entries}\vspace{1mm}
\begin{innercode}\vittag HR: 105--127 bpm (persistent tachycardia)\end{innercode}\end{responsebox}

% ======================================================================
% Event 24 -- llm_response  (6 parallel tool calls)
% ======================================================================
\begin{agentbox}
\agentheader
Let me search for specific adrenal-related conditions and check for
clinical notes that may have details about the adrenal insufficiency
etiology.
\end{agentbox}

% Event 25
\begin{toolbox}\toolheader{fhir\_condition\_search\_problems}\vspace{1mm}
\begin{innercode}\{"patient": "MRN4888657619", "code": "E24"\}\end{innercode}\end{toolbox}
\begin{responsebox}\respheader\hfill\textit{\small 0 entries}\end{responsebox}

% Event 26
\begin{toolbox}\toolheader{fhir\_condition\_search\_problems}\vspace{1mm}
\begin{innercode}\{"patient": "MRN4888657619", "code": "D35"\}\end{innercode}\end{toolbox}
\begin{responsebox}\respheader\hfill\textit{\small 0 entries}\end{responsebox}

% Event 27
\begin{toolbox}\toolheader{fhir\_condition\_search\_problems}\vspace{1mm}
\begin{innercode}\{"patient": "MRN4888657619", "code": "E23"\}\end{innercode}\end{toolbox}
\begin{responsebox}\respheader\hfill\textit{\small 0 entries}\end{responsebox}

% Event 28
\begin{toolbox}\toolheader{fhir\_condition\_search\_problems}\vspace{1mm}
\begin{innercode}\{"patient": "MRN4888657619", "code": "E03"\}\end{innercode}\end{toolbox}
\begin{responsebox}\respheader\hfill\textit{\small 0 entries}\end{responsebox}

% Event 29
\begin{toolbox}\toolheader{fhir\_observation\_search\_labs}\vspace{1mm}
\begin{innercode}\{"patient": "MRN4888657619", "code": "2141-0"\}\hspace{0.6em}\textit{\# ACTH}\end{innercode}\end{toolbox}
\begin{responsebox}\respheader\hfill\textit{\small 0 entries}\end{responsebox}

% Event 30 -- KEY TURN
\begin{toolbox}\toolheader{fhir\_document\_reference\_search\_clinical\_notes}\vspace{1mm}
\begin{innercode}\{"patient": "MRN4888657619", "date": "ge2022-06-01", "count": 10\}\end{innercode}\end{toolbox}
\begin{responsebox}\respheader\hfill\textit{\small 5 entries}\vspace{1mm}
\begin{innercode}\notetag 2022-06-02 Telephone Encounter (Endo)\end{innercode}
\vspace{1mm}
{\small\color{softtext}\textit{``Patient with known \textbf{secondary adrenal
insufficiency} diagnosed 2019 (morning cortisol \textbf{0.9 mcg/dL})
attributed to chronic exogenous ophthalmic corticosteroid drops and
opioid use. On hydrocortisone 15 mg/day. Now reports fatigue, BP
lability, HR 105\ldots''}}
\vspace{1mm}
\begin{innercode}\notetag 2022-06-04,-06,-08,-09 Additional endo notes\end{innercode}\end{responsebox}

% ======================================================================
% Event 31 -- llm_response  (5 parallel tool calls)
% ======================================================================
\begin{agentbox}
\agentheader
Excellent! The telephone encounter note is very informative. Let me now
search for additional specific data to complete the picture.
\end{agentbox}

% Event 32
\begin{toolbox}\toolheader{fhir\_document\_reference\_search\_clinical\_notes}\vspace{1mm}
\begin{innercode}\{"patient": "MRN4888657619", "date": "ge2022-06-08", "count": 10\}\end{innercode}\end{toolbox}
\begin{responsebox}\respheader\hfill\textit{\small 4 entries}\vspace{1mm}
\begin{innercode}\notetag Follow-up notes 2022-06-08 onward\end{innercode}\end{responsebox}

% Event 33
\begin{toolbox}\toolheader{fhir\_observation\_search\_labs}\vspace{1mm}
\begin{innercode}\{"patient": "MRN4888657619", "code": "3024-7"\}\end{innercode}\end{toolbox}
\begin{responsebox}\respheader\hfill\textit{\small 0 entries}\end{responsebox}

% Event 34
\begin{toolbox}\toolheader{fhir\_observation\_search\_labs}\vspace{1mm}
\begin{innercode}\{"patient": "MRN4888657619", "code": "11580-8"\}\end{innercode}\end{toolbox}
\begin{responsebox}\respheader\hfill\textit{\small 0 entries}\end{responsebox}

% Event 35
\begin{toolbox}\toolheader{fhir\_condition\_search\_problems}\vspace{1mm}
\begin{innercode}\{"patient": "MRN4888657619", "code": "E27.1"\}\end{innercode}\end{toolbox}
\begin{responsebox}\respheader\hfill\textit{\small 0 entries}\end{responsebox}

% Event 36
\begin{toolbox}\toolheader{fhir\_condition\_search\_problems}\vspace{1mm}
\begin{innercode}\{"patient": "MRN4888657619", "code": "E27.4"\}\end{innercode}\end{toolbox}
\begin{responsebox}\respheader\hfill\textit{\small 0 entries}\end{responsebox}

% ======================================================================
% Event 37 -- llm_response  (4 parallel tool calls)
% ======================================================================
\begin{agentbox}
\agentheader
Now let me search for the remaining telephone encounter notes and any
adrenal-specific diagnoses.
\end{agentbox}

% Event 38
\begin{toolbox}\toolheader{fhir\_document\_reference\_search\_clinical\_notes}\vspace{1mm}
\begin{innercode}\{"patient": "MRN4888657619", "date": "ge2022-06-12", "count": 10\}\end{innercode}\end{toolbox}
\begin{responsebox}\respheader\hfill\textit{\small 3 entries}\vspace{1mm}
\begin{innercode}\notetag 2022-06-12, -14, -17 Endo follow-up notes\end{innercode}\end{responsebox}

% Event 39
\begin{toolbox}\toolheader{fhir\_observation\_search\_labs}\vspace{1mm}
\begin{innercode}\{"patient": "MRN4888657619", "code": "1916-6"\}\end{innercode}\end{toolbox}
\begin{responsebox}\respheader\hfill\textit{\small 0 entries}\end{responsebox}

% Event 40
\begin{toolbox}\toolheader{fhir\_observation\_search\_labs}\vspace{1mm}
\begin{innercode}\{"patient": "MRN4888657619", "code": "3016-3"\}\end{innercode}\end{toolbox}
\begin{responsebox}\respheader\hfill\textit{\small 0 entries}\end{responsebox}

% Event 41
\begin{toolbox}\toolheader{fhir\_observation\_search\_labs}\vspace{1mm}
\begin{innercode}\{"patient": "MRN4888657619", "code": "1742-6"\}\end{innercode}\end{toolbox}
\begin{responsebox}\respheader\hfill\textit{\small 1 entry}\vspace{1mm}
\begin{innercode}\labtag ALT: 42.63 U/L (mildly elevated)\end{innercode}\end{responsebox}

% ======================================================================
% Event 42 -- llm_response  (2 tool calls)
% ======================================================================
\begin{agentbox}
\agentheader
Now let me search for the endocrinology clinic note from around 6/12
when the referral was completed.
\end{agentbox}

% Event 43
\begin{toolbox}\toolheader{fhir\_document\_reference\_search\_clinical\_notes}\vspace{1mm}
\begin{innercode}\{"patient": "MRN4888657619", "date": "le2022-05-23", "count": 10\}\end{innercode}\end{toolbox}
\begin{responsebox}\respheader\hfill\textit{\small 2 entries}\vspace{1mm}
\begin{innercode}\notetag 2022-05-23 Endocrinology clinic note\end{innercode}\end{responsebox}

% Event 44
\begin{toolbox}\toolheader{fhir\_observation\_search\_labs}\vspace{1mm}
\begin{innercode}\{"patient": "MRN4888657619", "code": "30000-4"\}\end{innercode}\end{toolbox}
\begin{responsebox}\respheader\hfill\textit{\small 0 entries}\end{responsebox}

% ======================================================================
% Event 45 -- llm_response  (1 tool call)
% ======================================================================
\begin{agentbox}
\agentheader
I now have comprehensive data. Let me compile all the clinical information
and write the management plan.
\end{agentbox}

% Event 46 -- write_file  (verbatim excerpt from agent-generated content)
\begin{toolbox}\toolheader{write\_file}\vspace{1mm}
\begin{innercode}\{"file\_path": "/workspace/output/management\_plan.txt",\\
\ "content": "\ldots\\
\ 7a. Etiology of Adrenal Insufficiency\\
\ This patient's adrenal insufficiency is\\
\ most consistent with \textbf{PRIMARY adrenal insufficiency}\\
\ \textbf{(Addison's disease)}. Supporting evidence includes:\\
\ \ \ - Elevated plasma renin activity (16.0 ng/mL/h)\\
\ \ \ - Low-normal sodium (135.1), high-normal potassium (4.95)\\
\ \ \ - Hypothyroidism, raising APS type 2 possibility\\
\ \\
\ 8b. Mineralocorticoid Replacement -- NEW ADDITION\\
\ \textbf{INITIATE FLUDROCORTISONE:}\\
\ \ \ - Fludrocortisone 0.1 mg PO daily (morning)\\
\ \\
\ 8c. Refer to Cardiology if BP remains poorly\\
\ \ \ \ \ controlled after 4--6 weeks\ldots"\}\end{innercode}\end{toolbox}
\begin{responsebox}\respheader\vspace{1mm}
\begin{innercode}\writtag \{"status": "ok", "bytes\_written": 14372,\\
\ \ "path": ".../management\_plan.txt"\}\end{innercode}\end{responsebox}

% ======================================================================
% Event 47 -- llm_response (no tool calls; final summary)
% ======================================================================
\begin{agentbox}
\agentheader
The comprehensive clinical assessment and management plan has been saved.
Let me provide a summary of the key findings and actions:

\textbf{\#\# Summary}

\textbf{\#\#\# Key Clinical Findings}
\begin{itemize}\setlength{\itemsep}{1pt}\leftskip-1em
\item \textbf{Patient}: 69-year-old female with \textbf{primary adrenal
  insufficiency} (supported by elevated renin activity 16.0 ng/mL/h,
  low-normal sodium 135.1, high-normal potassium 4.95), concurrent
  hypothyroidism (suggestive of \textbf{autoimmune polyglandular
  syndrome type 2})
\item \textbf{Current regimen}: Hydrocortisone 10 mg AM + 5 mg PM =
  \textbf{15 mg/day} (lower end of physiologic range)
\item \textbf{Symptoms}: Fatigue, blood pressure lability (118--164
  systolic), persistent tachycardia (105--127 bpm), nausea, vomiting,
  diarrhea, decreased appetite
\item \textbf{Critical gap}: No fludrocortisone prescribed despite
  primary AI with markedly elevated renin
\end{itemize}

\textbf{\#\#\# Management Plan}
\begin{enumerate}\setlength{\itemsep}{1pt}\leftskip-1em
\item \textbf{Hydrocortisone increase}: 10 $\rightarrow$ \textbf{15 mg AM},
  5 mg PM maintained $\rightarrow$ new total \textbf{20 mg/day}
\item \textbf{Initiate fludrocortisone 0.1 mg daily} --- essential for
  primary AI mineralocorticoid replacement
\item Cardiology referral if BP remains uncontrolled
\item Confirmatory labs: ACTH, 21-hydroxylase Ab, repeat renin/TSH
\item Sick-day stress dosing reinforcement
\end{enumerate}
\end{agentbox}

% ======================================================================
% Event 48 -- final_result (trajectory terminates)
% ======================================================================
\begin{agentbox}
\agentheader
\textit{Trajectory ended. Output file written; 44 tool calls total.}
\end{agentbox}